\pdfoutput=1

\documentclass[10pt,letterpaper]{article}

\usepackage{ncore,subfigure,times}

\usepackage{tikz}
\usetikzlibrary{trees}

\title{An Investigation Report on Auction Mechanism Design}
\author{Jinzhong Niu\\
Department of Computer Science\\
The Graduate School and University Center\\
The City University of New York\\
365, Fifth Avenue, New York, NY 10016\\
\texttt{jniu@gc.cuny.edu}
\and Simon Parsons\\
Department of Computer and Information Science\\
Brooklyn College\\
The City University of New York\\
2900 Bedford Avenue, Brooklyn, NY 11210\\
\texttt{parsons@sci.brooklyn.cuny.edu}
}

\begin{document}

%\enabletableofcontents

\maketitle

%\tableofcontents
%
%\newpage

%%%%%%%%%%%%%%%%%%%%%%%%%%%%%%%%%%%%%%%%%%%%%%%%%%%%%%%%%%%%%%%%%%%%%

\begin{abstract}
Auctions are markets with strict regulations governing the information available to traders in the market and the possible actions they can take. Since well designed auctions achieve desirable economic outcomes, they have been widely used in solving real-world optimization problems, and in structuring stock or futures exchanges. Auctions also provide a very valuable testing-ground for economic theory, and they play an important role in computer-based control systems.

Auction mechanism design aims to manipulate the rules of an auction in order to achieve specific goals. Economists traditionally use mathematical methods, mainly game theory, to analyze auctions and design new auction forms. However, due to the high complexity of auctions, the mathematical models are typically simplified to obtain results, and this makes it difficult to apply results derived from such models to market environments in the real world. As a result, researchers are turning to empirical approaches.

This report aims to survey the theoretical and empirical approaches to designing auction mechanisms and trading strategies with more weights on empirical ones, and build the foundation for further research in the field.

\end{abstract}

%
%\newpage
%

%%%%%%%%%%%%%%%%%%%%%%%%%%%%%%%%%%%%%%%%%%%%%%%%%%%%%%%%%%%%%%%%%%%%%%
\section{Auctions}
\label{sec:auctions}
%%%%%%%%%%%%%%%%%%%%%%%%%%%%%%%%%%%%%%%%%%%%%%%%%%%%%%%%%%%%%%%%%%%%%%

%---------------------------------------------------------------------
\subsection{Auction types}
\label{sec:auctions:types}
%---------------------------------------------------------------------

A \emph{market} is a set of arrangements by which buyers and sellers, collectively known as \emph{traders}, are in contact to exchange goods or services. \emph{Auctions}, a subclass of markets with strict regulations governing the information available to traders in the market and the possible actions they can take, have been widely used in solving real-world optimization problems, and in structuring stock or futures exchanges.

The most common kind of auction is the \emph{English auction}, in which there is a single seller, and multiple buyers compete by making increasing bids for the commodity (good or service) being auctioned; the one who offers the highest price wins the right to purchase the commodity. Since only one type of trader---buyers---makes offers in an English auction, the auction belongs to the class of \emph{single-sided auctions}. Another common single-sided auction is the \emph{Dutch auction}, in which the auctioneer initially calls out a high price and then gradually lowers it until one bidder indicates they will accept that price.

Another class of single-sided auctions is the class of \emph{sealed-bid auctions}, in which all buyers submit a single bid and do so simultaneously, i.e., without observing the bids of the others or if the others have bid. Two common sealed-bid auctions are the \emph{first-price auction} and the \emph{second-price auction} or \emph{Vickrey auction} \cite{vickrey61@counterspeculation}. In both types of sealed-bid auctions, the highest bidder obtains the commodity. In the former, the highest bidder pays the price they bid, while in the latter, they pay the second highest price that was bid.

These four single-sided auctions---English, Dutch, first-price sealed-bid, and Vickrey---are commonly referred to as the \emph{standard auctions} and were the basis of much early research on auctions.

In addition, there are \emph{double-sided auctions} or \acro{da}s\footnote{The terminology is not standardized, and sometimes these are called \emph{bid-ask auctions}. Note that \cite{friedman93@double-auction-institution-survey} used the term ``double auction" to refer to what we call a continuous double auction in this report.}, in which both sellers and buyers make offers, or \emph{shouts}. The two most common forms of \textsc{da} are \emph{clearing houses} or \textsc{ch}s\footnote{These are sometimes called \emph{call markets} or \emph{static double auctions}.} and \emph{continuous double auctions} or \textsc{cda}s. In a \textsc{ch}, an auctioneer first collects \emph{bids}---shouts from buyers---and \emph{asks}---shouts from sellers, and then clears the market at a price where the quantity of the commodity supplied equals the quantity demanded. This type of market clearing guarantees that if a given trader is involved in a transaction, all traders with more competitive offers are also involved.\footnote{That is only \emph{intra-marginal} traders are involved in transactions.} In a \textsc{cda}, a trader can make a shout and accept an offer from someone at any time. This design makes a \textsc{cda} able to process many transactions in a short time, but permits extra-marginal traders to make deals. Both kinds of \acro{da} are of practical importance, with, for example, \acro{cda} variants being widely used in real-world stock or trading markets including the New York Stock Exchange (\acro{nyse}) and the Chicago Mercantile Exchange (\acro{cme}).

In some auctions, traders can place shouts on combinations of items, or ``packages", rather than just individual items. They are called \emph{combinatorial auctions}. A common procedure in these markets is to auction the individual items and then at the end to accept bids for packages of items. Combinatorial auctions present a host of new challenges as compared to traditional auctions, including the so-called \emph{winner determination problem}, which is how to efficiently determine the allocation once the bids have been submitted to the auctioneer.

Traders, in some cases, are allowed to both sell and buy during an auction. Such traders are called \emph{two-way traders}, while those that only buy or only sell are called \emph{one-way traders}.

This report will mainly discuss non-combinatorial \acro{da}s, especially \acro{cda}s, populated by one-way traders.

%---------------------------------------------------------------------
\subsection{Supply, demand and equilibrium}
\label{sec:auctions:supply-demand-equilibrium}
%---------------------------------------------------------------------

\begin{figure*}[!tb]
  \begin{center}
  \mbox{
    \subfigure[Underlying]{\label{fig:underlying-supply-demand}\includegraphics[scale=0.35]{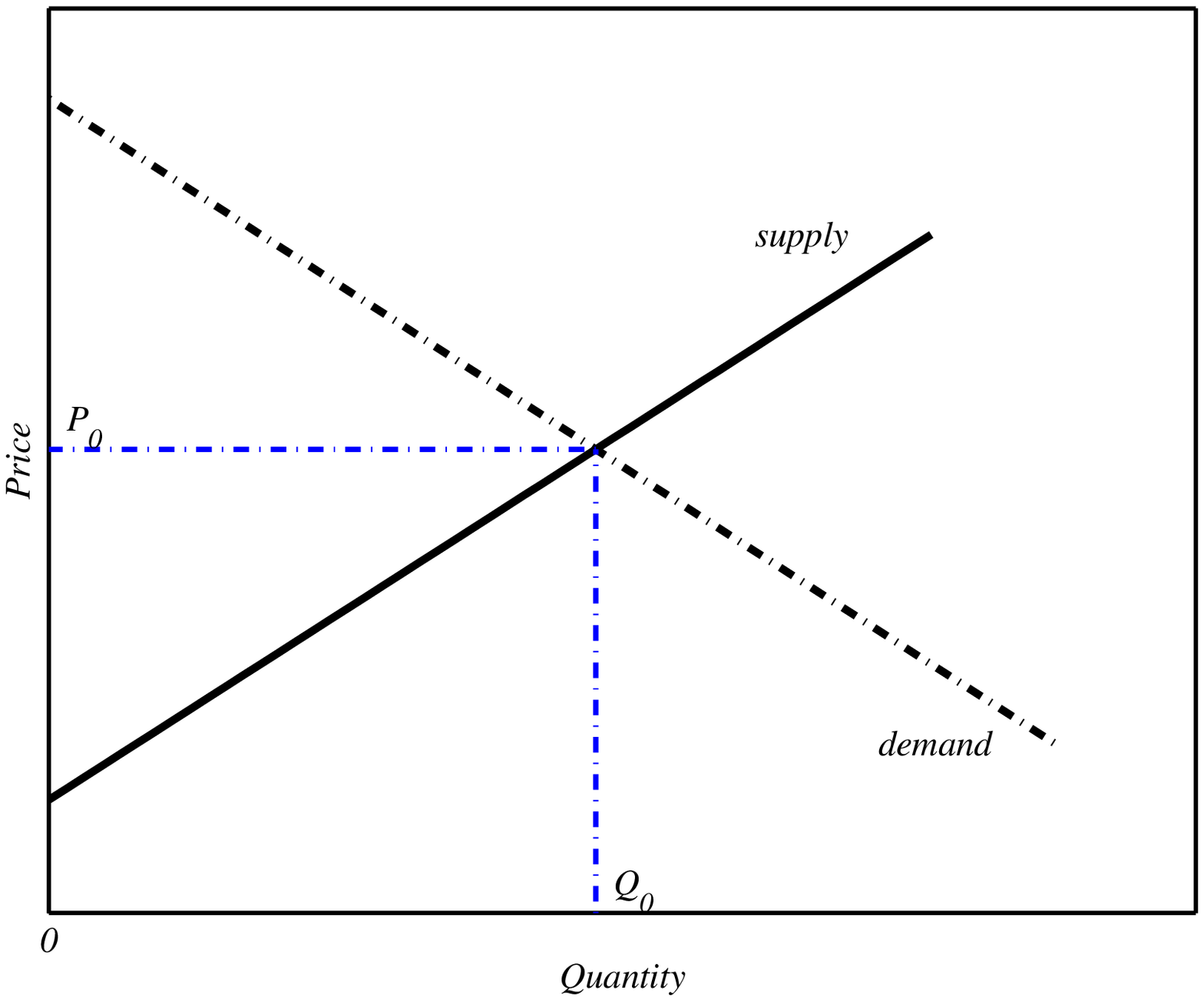}}
    \subfigure[Apparent]{\label{fig:apparent-supply-demand}\includegraphics[scale=0.35]{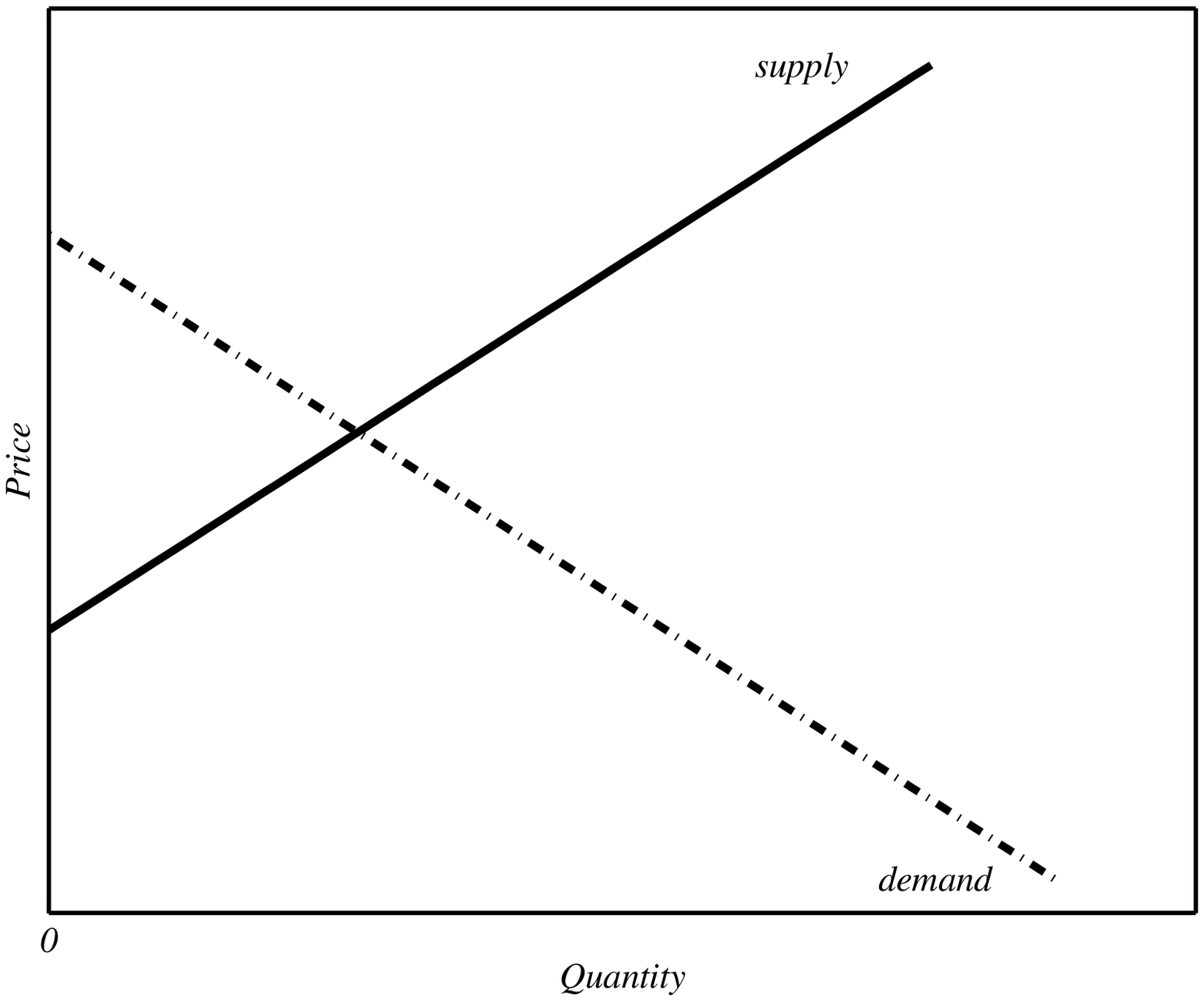}}
  }
  \caption{Typical supply and demand curves.}
  \label{fig:supply-demand}
  \end{center}
\end{figure*}

A central concern in studies of auction mechanisms are the supply and demand schedules in a market. The quantity of a commodity that buyers are prepared to purchase at each possible price is referred to as the \emph{demand}, and the quantity of a commodity that sellers are prepared to sell at each possible price is referred to as the \emph{supply}. Thus if $price$ is plotted as a function of $quantity$, the \emph{demand curve} slopes downward and the \emph{supply curve} slopes upward, as shown in Figure~\ref{fig:underlying-supply-demand}, since the greater the price of a commodity, the more sellers are inclined to sell and the fewer buyers are willing to buy. Typically, there is some price at which the quantity demanded is equal to the quantity supplied. Graphically, this is the intersection of the demand and supply curves. The price is called the \emph{equilibrium price}, and the corresponding quantity of commodity that is traded is called the \emph{equilibrium quantity}. The equilibrium price and equilibrium quantity are denoted as $P_0$ and $Q_0$ respectively in Figure~\ref{fig:underlying-supply-demand}.

Each trader in an auction presumably has a limit price, called its \emph{private value}, below which sellers will not sell and above which buyers will not buy. The private values of traders are not publicly known in most practical scenarios. What is known instead are the prices that traders offer. Self-interested sellers will presumably offer higher prices than their private values to make a profit and self-interested buyers tend to offer lower prices than their private values to save money. The prices and quantities that are offered also make a set of supply and demand curves, called the \emph{apparent supply and demand curves}, while the curves based on traders' private values are called the \emph{underlying supply and demand}.\footnote{Following the terminology in \cite{cliff97@minimal-intell-agent}.} Figure~\ref{fig:apparent-supply-demand} shows that the apparent supply curve shifts up compared to the underlying supply curve in Figure~\ref{fig:underlying-supply-demand}, while the apparent demand curve shifts down.

When traders are excessively greedy, the apparent supply and demand curves do not intersect and thus no transactions can be made between sellers and buyers unless they compromise on their profit levels and adjust their offered prices.

%---------------------------------------------------------------------
\subsection{A typical time series of shouts}
\label{sec:auctions:time-series}
%---------------------------------------------------------------------

\begin{figure*}[!tb]
  \begin{center}
  \mbox{
    \subfigure{\includegraphics[scale=0.6]{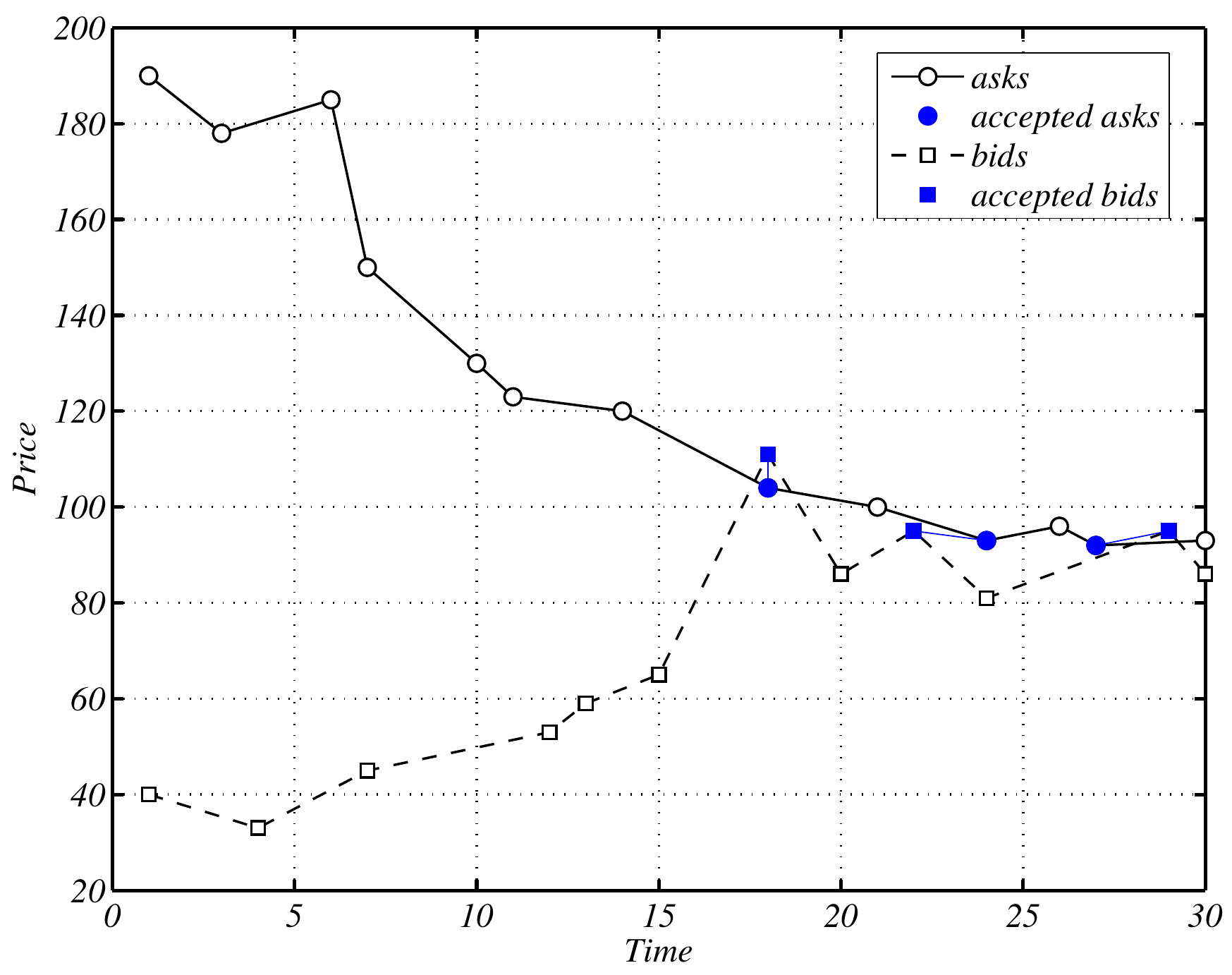}}
  }
  \caption{Time series of asks and bids.}\label{fig:time-series}
  \end{center}
\end{figure*}

In a \acro{cda}, buyers and sellers not only `haggle' on prices in a collective manner, but they also face competition from opponents on the same side of the market. Thus buyers, for example, are not only collectively trying to drive prices down, against the wishes of sellers, but they are also individually trying to ensure that they, rather than other buyers, make profitable trades. This leads to shouts becoming more and more competitive over time in a given market. Figure~\ref{fig:time-series} shows a typical time series of shouts in a \acro{da}. Ask prices usually start high while bid prices start low. Gradually, traders adjust their offered prices, or make new shouts, closing the gap between standing asks and bids until the price of a bid surpasses that of an ask. Such an overlap results in a transaction, shown as a solid bar between the matched ask and bid in Figure~\ref{fig:time-series}.

In the market depicted in Figure~\ref{fig:time-series}, newly placed bids (asks) do not have to beat the outstanding bids (asks). However in some variants of the \acro{cda} including the market operated by the \acro{nyse}, new shouts must improve on existing ones. This requirement is commonly referred to as the \emph{\acro{nyse} shout improvement rule} \cite{easley93price-formation}.

In some real-world stock markets, including the \acro{nyse} and the \acro{nasdaq} markets, trades are made through \emph{specialists} or \emph{market makers}, who buy or sell stock from their own inventory to keep the market liquid or to prevent rapid price changes.\footnote{In the \acro{nyse}, a given stock is traded through a single specialist, and in the \acro{nasdaq}, a stock may be dealt with by multiple competing market makers.} Each specialist is required to publish on a regular and continuous basis both a \emph{bid quote}, the highest price it will pay a trader to purchase securities, and an \emph{ask quote}, the lowest price it will accept from a trader to sell securities. The specialist is obligated to stand ready to buy at the bid quote or sell at the ask quote up to a certain number of shares. The range between the lower bid quote and the higher ask quote is called the \emph{bid-ask spread}, which, according to stock exchange regulations, must be suitably small. If buy orders temporarily outpace sell orders, or conversely if sell orders outpace buy orders, the specialist is required to use its own capital to minimize the imbalance. This is done by buying or selling against the trend of the market until a price is reached at which public supply and demand are once again in balance. Maintaining a bid-ask spread creates risk for a specialist, but when well maintained, also brings huge profits, especially in an active market \cite{bao01market-making}.

Markets involving specialists that present quotes are called \emph{quote-driven markets}. Another class of markets are \emph{order-driven markets}, in which all of the orders of buyers and sellers are displayed. This contrasts with quote-driven markets where only the orders of market makers are shown. An example of an order-driven market is the market formed by \emph{electronic communication networks} or \acro{ecn}s. These are electronic systems connecting individual traders so that they can trade directly between themselves without having to go through a middleman like a market maker. The biggest advantage of this market type is its transparency. The drawback is that in an order-driven market, there is no guarantee of order execution, meaning that a trader has no guarantee of making a trade at a given price, while it is guaranteed in a quote-driven market. There are markets that combine attributes from quote- and order-driven markets to form hybrid systems.

Our discussion above may give the impression that in real markets trade orders are made directly by the individuals who want to buy or sell stock. In practice, traders commonly place orders through brokerage firms, which then manage the process of executing the orders through a market.\footnote{\url{http://www.sec.gov/investor/pubs/tradexec.htm}\ gives a detailed illustration of how a trade order is executed through a brokerage firm.}

%---------------------------------------------------------------------
\subsection{Performance metrics}
\label{sec:auctions:performance-metrics}
%---------------------------------------------------------------------

Auctions with different rules and populated by different sets of traders may vary greatly in performance. Popular performance measurements include, but are not limited to, \emph{allocative efficiency} and the \emph{coefficient of convergence}.

%---------------------------------------------------------------------
\subsubsection{Allocative efficiency}
\label{sec:auctions:performance-metrics:eff}
%---------------------------------------------------------------------

The allocative efficiency of an auction, denoted as $E_a$, is used to measure how much social welfare is obtained through the auction. The \emph{actual overall profit}, $P_a$, of an auction is:
\begin{equation}
\label{equ:actual-profit}
P_{a}=\sum_{i}|v_{i}-p_{i}|
\end{equation}
where $p_i$ is the transaction price of a trade completed by agent $i$ and $v_i$ is the private value of agent $i$, where $i$ ranges over all agents who trade. The \emph{theoretical} or \emph{equilibrium profit}, $P_e$, of an auction is:
\begin{equation}
\label{equ:theoretical-profit}
P_{e}=\sum_{i}|v_{i}-p_{0}|
\end{equation}
for all agents whose private value is no less competitive than the equilibrium price, where $p_{0}$ is the equilibrium price. Given these:
\begin{equation}
\label{equ:efficiency}
E_{a}=\frac{100{P_{a}}}{P_{e}}
\end{equation}
$E_a$ is thus a measure of the proportion of the theoretical profit that is achieved in practice.

%---------------------------------------------------------------------
\subsubsection{Convergence coefficient}
\label{sec:auctions:performance-metrics:conv-coeff}
%---------------------------------------------------------------------

The convergence coefficient, denoted as $\alpha$, was introduced by Smith \cite{smith62@competitive-market-behavior} to measure how far an active auction is away from the equilibrium point. It actually measures the relative \acro{rms} deviation of transaction prices from the equilibrium price:
\begin{equation}
\label{equ:alpha}
\alpha=\frac{100}{p_{0}}{{\sqrt{\frac{1}{n}\sum_{i=1}^{n}{(p_{i}-p_{0})^{2}}}}}
\end{equation}
Since markets with human traders often trade close to the equilibrium price, $\alpha$ is used as a way of telling how closely artificial traders approach human trading performance.

%\clearpage

%%%%%%%%%%%%%%%%%%%%%%%%%%%%%%%%%%%%%%%%%%%%%%%%%%%%%%%%%%%%%%%%%%%%%%
\section{Game theory}
\label{sec:game-theory}
%%%%%%%%%%%%%%%%%%%%%%%%%%%%%%%%%%%%%%%%%%%%%%%%%%%%%%%%%%%%%%%%%%%%%%

Research on auctions originally interested mathematical economists. They view auctions as games and have successfully applied traditional analytic methods from game theory. This section therefore takes an overlook at basic concepts in game theory.

%---------------------------------------------------------------------
\subsection{Games}
\label{sec:game-theory:game}
%---------------------------------------------------------------------

The games studied by game theory are well-defined mathematical objects. A game is usually represented in its \emph{normal form}, or \emph{strategic form}, which is a tuple $$\langle \A_1, \cdots, \A_n, R_1, \cdots, R_n\rangle.$$ $n$ is the number of players, $\A_i$ is the set of \emph{actions} available to player $i$, and $R_i$ is the \emph{payoff} or \emph{utility function} $\A\rightarrow \Re$, where $\A$ is the joint action space $\A_1\times \cdots \times \A_n$.

When a player needs to act, it may follow a \emph{pure strategy}, choosing an action, $a_i$, from its action set, or a \emph{mixed strategy}, $\pi_i$, choosing actions according to a probability distribution. The strategy set of player $i$, denoted as $\Pi_i$, is the same thing as a set of probability distributions over $\A_i$, denoted as $\Delta(\A_i)$. A joint strategy for all players is called a \emph{strategy profile}, denoted as $\pi$, and $\pi(a)$ is the probability all players choose the joint action $a$ from $\A$. Thus player $i$'s payoff for the strategy profile $\pi$ is: $$R_i(\pi)=\sum_{a\in \A}{\pi(a) \ R_i(a)}.$$ In addition, $\Pi$ denotes the set of all possible strategy profiles,\footnote{It can also be represented as $\Delta(\A_1)\times \cdots \times \Delta(\A_n)$.} $\pi_{-i}$ is a strategy profile for all players except $i$, and $\langle \pi_i,\pi_{-i}\rangle$ is the strategy profile where player $i$ uses strategy $\pi_i$ and the others use $\pi_{-i}$.

\begin{figure*}[!tb]
  \begin{center}
  \begin{tikzpicture}[grow=east]
  \tikzstyle{level 1}=[level distance=2cm, sibling distance=2cm]
  \tikzstyle{level 2}=[level distance=2cm, sibling distance=1cm]
  \tikzstyle{tree}=[fill,circle,minimum size=1pt]
  \tikzstyle{terminal}=[right=5pt]
  \tikzstyle{annotation}=[text centered]
  \node[tree] {}
    child {node[tree] {}
      child {node[tree] {}
        node[terminal] {$0,0$}
        edge from parent
        node[annotation,below] {$a'_2$}
      }
      child {node[tree] {}
        node[terminal] {$2,1$}
        edge from parent
        node[annotation,above] {$a_2$}
      }
      edge from parent
      node[annotation,below] {$a'_1$}
    }
    child {node[tree] {}
      child {node[tree] {}
        node[terminal] {$1,2$}
        edge from parent
        node[annotation,below] {$a'_2$}
      }
      child {node[tree] {}
        node[terminal] {$3,3$}
        edge from parent
        node[annotation,above] {$a_2$}
      }
      edge from parent
      node[annotation,above] {$a_1$}
    };
  \end{tikzpicture}
  \caption{A game represented in extensive form.}
  \label{fig:extensive-form-game}
  \end{center}
\end{figure*}
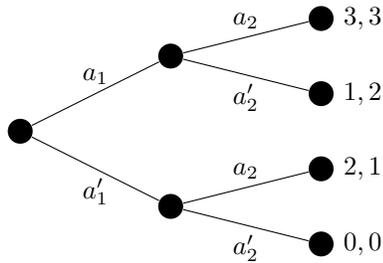

A normal-form game is typically illustrated as a matrix with each dimension listing the choices of one player and each cell containing the payoffs of players for the corresponding joint action. Figure~\ref{fig:normal-form-prisoner-dilemma} shows the normal form of the well-known Prisoner's Dilemma game. Alternatively, games may be represented in \emph{extensive form}, which is a tree, as in Figure~\ref{fig:extensive-form-game}. The tree starts with an initial node and each node represents a state during play. At each non-terminal node, a given player has the choice of action. Different choices lead to different child nodes, until a terminal node is reached where the game is complete and the payoffs to players are given.

\begin{figure*}[!tb]
  \begin{center}
\[
      \begin{tabular}{|r|c|c|}
      \hline
                & \textbf{cooperate} & \textbf{defect} \\ \hline
      \textbf{cooperate} & 3,\ 3 & 0,\ 4 \\ \hline
      \textbf{defect} & 4,\ 0 & 1,\ 1 \\ \hline
      \end{tabular}
\]
  \caption{The payoff matrix of the Prisoner's Dilemma game.}
  \label{fig:normal-form-prisoner-dilemma}
  \end{center}
\end{figure*}

A game may be \emph{cooperative} or \emph{noncooperative}, as players in these games are respectively \emph{cooperative} or \emph{self-interested}. Cooperative players share a common payoff function, i.e., $$\forall i,j\ \ R_i=R_j,$$ whereas self-interested players typically have distinct payoff functions. In both cases, players need to coordinate in a certain way to `assist' each other in achieving their goals\footnote{The goal of the game designer is also an issue in many situations. Therefore in some parts of the literature, games are considered cooperative as long as they produce a desired systematic outcome even with self-interested players.} \cite{lesser99coop-mas}. If the payoffs of all players for each strategy profile sum to zero, the noncooperative game is called a \emph{zero-sum} game, i.e., $$\forall \pi\in \Pi,\;\ \sum_{i=1,\cdots,n}{R_i(\pi)=0}.$$ Zero-sum games are a special case of a more general class of games called \emph{constant-sum} games, where the sum of all payoffs for each outcome is a constant but may not necessarily be zero. Non-zero-sum games are sometimes referred to as \emph{general-sum} games. In economic situations, the exchange of commodities is considered general-sum, since both parties gain more through the transaction than if they had not transacted (otherwise the exchange would not have happened, assuming both are rational).

In some games, the payoffs for playing a particular strategy remain unchanged as long as the other strategies employed collectively by the players are same, no matter which player takes which action. These games are called \emph{symmetric games} and the rest are \emph{asymmetric games}. For example, the Prisoner's Dilemma game given above is symmetric.

Players may take actions \emph{simultaneously} or \emph{sequentially}. In a sequential game, players have alternating turns to take actions and a player has knowledge about what actions the other players have taken previously. Simultaneous games are usually represented in normal form, and sequential games are usually represented in extensive form. A sequential game is considered a game of \emph{perfect information} if all players know all the actions previously taken by the other players. A similar concept is a game of \emph{complete information}, which means all players in the game know the strategies and payoff functions of the other players. In some sense, complete information may be viewed as capturing static information about a game while perfect information addresses dynamic information that becomes available during runs (or instances) of the game.

%---------------------------------------------------------------------
\subsection{Nash equilibrium}
\label{sec:game-theory:nash-eq}
%---------------------------------------------------------------------

There are various solutions to a normal-form game depending upon the properties of the game and preferences over outcomes.

A strategy, $\pi_i$, is said to be \emph{dominant} if it always results in higher payoffs than any other choice no matter what the opponents do, i.e., $$\forall \pi'_i\in \Pi,\;\ R_i(\langle \pi_i,\pi_{-i}\rangle)\geq R_i(\langle \pi'_i,\pi_{-i}\rangle).$$ In the example of the Prisoner's Dilemma game, the choice \textbf{defect} dominates \textbf{cooperate} for either player, though ironically both will be better off if they choose to cooperate and know that the other will also.

In many games there are, however, no dominant strategies. To conservatively guarantee the best worst-case outcomes, a player may play the \emph{minimax} strategy, which is $$\argmax_{\pi_i\in \Pi_i}\min_{a_{-i}\in \A_{-i}}{R_i(\langle \pi_i, \pi_{-i}\rangle)},$$ where $a_{-i}$ and $\A_{-i}$ are respectively a joint action for all players except $i$ and the set of joint actions for them. In theory, this can be solved via linear programming, but clearly there are many games that are too large to be solved in practice.

Another approach to solving the problem is to find the \emph{best response} strategies to the strategies of the other players. These can be defined as $$BR_i(\pi_{-i})=\{\pi_i | \forall \pi'_i, \;\ R_i(\langle \pi_i,\pi_{-i}\rangle)\geq R_i(\langle \pi'_i,\pi_{-i}\rangle)\}.$$  A joint strategy forms a \emph{Nash equilibrium} or \acro{ne} if each individual strategy is the best response to the others' strategies. When a \acro{ne} is reached, no player can be better off unilaterally, given that the other players stay with their strategies. In the example of the Prisoner's Dilemma game, \textbf{$\langle$defect,\ defect$\rangle$} is a \acro{ne}.

Although Nash \cite{nash50ne} showed that all finite normal-form games have at least one \acro{ne}, Nash equilibria are generally difficult to achieve. On the one hand, Conitzer and Sandholm \cite{conitzer02ne-complexity} proved that computing Nash equilibria is likely \acro{np}-hard; on the other hand, some games involve more than one \acro{ne}, thus without some extra coordination mechanism, no player knows which equilibrium the others would choose.

Many papers have been concerned with ``equilibrium refinements" so as to make one equilibrium more plausible than another, however it seems to lead to overly complicated models that are difficult to solve. A more practical approach is to allow players to learn by playing a game repeatedly. A \emph{repeated game} is a game made up from iterations of a single normal-form game, in which a player's strategy depends upon not only the one-time payoffs of different actions but also the history of actions taken by its opponents in preceding rounds. Such a game can be viewed as a system with multiple players and a single state, since the game setting does not change across iterations. If the setting changes over time, the game becomes a \emph{stochastic game}. A stochastic game involves multiple states and the player payoff functions relate to both their actions in each interaction and the current state. The goal of a player in such a game is to maximize its long-term return, which is sometimes defined as the average of all one-time payoffs or the discounted sum of those payoffs.

Brown \cite{brown51fictitious-play} introduced a learning method, called \emph{fictitious play}, for games in which all the other players use stationary strategies. With this method, the player in question keeps a record of how many times the other players have taken each action and uses the frequencies of actions to estimate the probabilities of actions in an opponent's strategy. Then the player chooses a best-response strategy based on its belief. If the player's belief converges, what it converges to and its own best-response strategy form a \acro{ne} \cite{fudenberg:levine98learning-in-games}. The method becomes flawed if players adopt non-stationary strategies and in some games the belief simply does not converge \cite{wikipedia-site}.

Another promising approach is to analyze the situation with evolutionary methods. This approach assumes there is a large population of individuals and each strategy is played by a certain fraction of these individuals. Then, given the distribution of strategies, individuals with better average payoffs will be more successful than others, so that their proportion in the population will increase over time. This, in turn, may affect which strategies are better than others. In many cases, the dynamic process will move to an equilibrium. The final result, which of possibly many equilibria the system achieves, will depend on the initial distribution. The evolutionary, population-dynamic view of games is useful because it does not require the assumption that all players are sophisticated and think the others are also rational, an assumption that is often unrealistic. Instead, the notion of \emph{rationality} is replaced with the much weaker concept of \emph{reproductive success}.

A related concept, considering the overall outcome rather than individual payoffs, is \emph{Pareto optimality}. A strategy profile, $\pi^*$, is \emph{Pareto optimal}, or \emph{Pareto efficient}, if there exists no other strategy profile producing higher payoffs for all players, i.e., $$\forall \pi\in \Pi,\ \exists i (R_i(\pi)>R_i(\pi^*)) \Rightarrow \exists j (R_j(\pi^*)>R_j(\pi)).$$ In the Prisoner's Dilemma game, all pure strategy profiles except for \textbf{$\langle$defect,\ defect$\rangle$} are Pareto optimal. A Pareto optimal outcome is highly desirable, but usually difficult to achieve. Self-interested players tend to take locally optimal actions that may not collectively be Pareto optimal. In the Prisoner's Dilemma game, \textbf{$\langle$cooperate,\ cooperate$\rangle$} instead of the \acro{ne} \textbf{$\langle$defect,\ defect$\rangle$} obviously causes both players to be better off.

In games of incomplete information, to utilize the concept of \acro{ne}, each player needs to maintain an estimate of the others' strategies so as to come up with a best-response strategy, where Bayes' theorem is used to update or revise beliefs following interactions with opponents. The concept of equilibrium therefore becomes \emph{Bayesian Nash Equilibrium}, or \acro{bne}. That is each player's strategy is a function of her own information, and maximizes her expected payoff given other players' strategies and given her beliefs about other players' information \cite{klemperer99auction-theory, wikipedia-site}.

%\clearpage

%%%%%%%%%%%%%%%%%%%%%%%%%%%%%%%%%%%%%%%%%%%%%%%%%%%%%%%%%%%%%%%%%%%%%%
\section{Auction theory}
\label{sec:auction-theory}
%%%%%%%%%%%%%%%%%%%%%%%%%%%%%%%%%%%%%%%%%%%%%%%%%%%%%%%%%%%%%%%%%%%%%%

Auctions are a way to enable interactions among traders, and traders make profits as a result of transactions. Vickrey \cite{vickrey61@counterspeculation} pioneered the approach of thinking about a market institution as a game of incomplete information since traders do not know each others' private values.

In research on single-sided auctions, the main goal is to find mechanisms that maximize the profit of sellers,\footnote{There is no formal distinction between \emph{normal auctions}, in which the auctioneer is the seller and the participants are buyers who have values for the object(s) to be sold, and \emph{procurement auctions}, where the auctioneer is a buyer and the participants are sellers who have costs of supplying the object(s) to be bought.} who are special players in the auctioning games. while in double-sided auctions, research focuses on maximizing social welfare and identifying how price formation develops dynamically.

%---------------------------------------------------------------------
\subsection{Revenue equivalence theorem}
\label{sec:auction-theory:revenue-equivalence}
%---------------------------------------------------------------------

By assuming a fixed number of ``symmetric",\footnote{That is bidders' private values are drawn from a common distribution.} risk-neutral\footnote{In economics, the term \emph{risk neutral} is used to describe an individual who cares only about the expected return of an action, and not the \emph{risk} (variance of outcomes or the potential gains or losses). A risk-neutral person will neither pay to avoid risk nor actively take risks. Similarly, there are \emph{risk-averse} and \emph{risk-seeking} individuals; they respectively favor the (usually lower) outcome with more certainty and the highest possible outcome (usually with lower probability) \cite{wikipedia-site}.} bidders, who each want a single unit of goods, have a private value for the object, and bid independently, Vickrey showed that the seller can expect equal profits on average from all the standard types of auctions. This finding is called the \emph{Revenue Equivalence Theorem}. This theorem provides the foundation for the analysis of \emph{optimal auctions}\footnote{Auctions that maximize the expected profit of sellers.} and much subsequent research can be understood in terms of this theorem. Numerous articles have reported how its results are affected by relaxing the assumptions behind it.

The assumption that each trader knows the value of the goods being traded, and that these values are all private and independent of each other is commonly called the \emph{private-value model} \cite{klemperer99auction-theory, parsons:klein04bluffer-guide2auction}.

In some cases, by contrast, the actual value of the goods is the same for everyone, but bidders have different private information about what that value actually is. In these cases, a bidder will change her estimate of the value if she learns another bidder's estimate, in contrast to the private-value case in which her value would be unaffected by learning any other bidder's preferences or information. This is called the \emph{pure common-value model} \cite{wilson69disparate-information}. The winner in this scenario is the individual who makes the highest estimate of the value, and this tends to be an overestimate of the value. This overestimation is called the \emph{winner's curse}. If all the bidders have the existence of the winner's curse in mind, the highest bid in first-price auctions tends to be lower than in those second-price auctions, though it still holds that the four standard auctions are revenue-equivalent.

A general model encompassing both the private-value model and the pure common-value model as special cases is the \emph{correlated-value model} \cite{milgrom:weber82theory-of-auctions}. This assumes that each bidder receives a private information signal, but allows each bidder's value to be a general function of \emph{all} the signals.\footnote{That is, bidder $i$ receives signal $t_i$ and would have value $v_i(t_1,\ldots,t_n)$ if all bidders' signals were available to her. In the private-value model $v_i(t_1,\ldots,t_n)$ is a function only of $t_i$. In the pure common-value model $v_i(t_1,\ldots,t_n)=v_j(t_1,\ldots,t_n)$ for all $i$ and $j$.} Milgrom and Weber analyzed auctions in which bidders have \emph{affiliated information},\footnote{Roughly speaking, bidders' information is affiliated if when one bidder has more optimistic information about the value of the prize, then it is more likely that other bidders' information will also be optimistic.} and showed that the most profitable standard auction is then the ascending auction.

Myerson \cite{myerson81optimal} demonstrated how to derive optimal auctions when the assumption of symmetry fails. Maskin and Riley \cite{maskin:riley84risk-averse} considered the case of risk-averse bidders, in which case the first-price sealed-bid auction is the most profitable of the standard auctions.

For practical reasons, it is more important to remove the assumptions that the number of bidders is unaffected by the auction design, and that the bidders necessarily bid independently of each other. According to \cite{klemperer99auction-theory}, sealed-bid designs frequently (but not always) both attract a large number of serious bidders and are better at discouraging collusion than English auctions.

%---------------------------------------------------------------------
\subsection{On double-sided auctions}
\label{sec:auction-theory:da}
%---------------------------------------------------------------------

In contrast with simple single-sided auctions, where the goals of auction mechanism designers reflect the interests of the single seller, double-sided auctions aim to maximize the collective interests of all traders, or in other words, the social welfare, i.e., the total surplus all traders earn in an auction. Numerous publications have reported theoretical assertions or empirical observations of high efficiency in a variety of double-sided auctions, and have discussed what leads to the maximization of social welfare.

Chatterjee and Samuelson \cite{chatterjee:samuelson83bargaining} made the first attempt to analyze double auctions considering a special case of the \acro{ch} involving a single buyer and a single seller. In this auction, the transaction price is set at the midpoint of the interval of market-clearing prices when the interval is non-empty. They found linear \acro{bne} bidding strategies which miss potential transactions with probability of 1/6. Satterthwaite and Williams \cite{satterthwaite89rate-convergence} analyzed a generalized version of this auction---so-called \emph{$k$-double auction} or $k$-\acro{da}---which involves $m$ sellers and $m$ buyers, and sets the transaction price at other points in the interval of market-clearing prices. They showed that in \acro{bne}s the differences between buyers' bids and true values are $O(1/m)$ and foregone gains\footnote{Foregone gain means the missed profit compared with the profit that would have been made if the market cleared at the equilibrium price.} from trade are $O(1/m^2)$, so \emph{ex post}\footnote{This refers to the value that is actually observed or the value calculated after an event occurs, in contrast to \emph{ex ante}, which means the expected value calculated before the resolution of uncertainty.} inefficiency vanishes reasonably fast as the market gets larger.

Wilson \cite{wilson87equilibrium} first studied the generalization of games of incomplete information to \acro{cda}s, in particular, \acro{cda}s in which each agent can trade at most one indivisible unit and, given the bids and asks, the maximum number of feasible trades are made at a price a fraction $k$ of the distance between the lowest and highest feasible market clearing prices. He proposed a strategy for buyers and sellers in which a trader waits for a while before making bids or asks. Then the trader conducts a Dutch auction until an offer from the other side is acceptable. This strategy produces a nearly \emph{ex post} efficient final allocation.

Wurman \textit{et al.} \cite{wurman98@flexible-double-auction-4-ecommerce} carried out an incentive compatibility analysis on a \acro{ch} which is assumed to have $M$ bids and $N$ asks. They showed that the ($M+1$)st-price (or $N$th-lowest-price) clearing policy is incentive compatible for single-unit buyers under the private-value model, as is the $M$th-price (or ($N+1$)st-lowest-price) auction for sellers. The only way to get incentive compatibility for both buyers and sellers is for some party to subsidize the auction. Myerson and Satterthwaite \cite{myserson:satterthwaite83efficiency-mechanism} showed that there does not exist any bargaining mechanism that is individually rational, efficient, and Bayesian incentive compatible for both buyers and sellers without any outside subsidies.

As Friedman \cite{friedman93@double-auction-institution-survey} pointed out, though theoretically it is natural to model \acro{da}s as a game of incomplete information, the assumption of prior common knowledge in the incomplete information approach may not hold in continuous auctions or may involve incredible computational complexity. This is because at every moment, a trader needs to compute expected utility-maximizing shouts based on the shout and transaction history of the auction and the length of time the auction has to go. On the other hand, laboratory results have shown that \acro{da} outcomes are quite insensitive to the number of traders beyond a minimal two or three active buyers and two or three active sellers.\footnote{This observation and the above result of $O(1/m^2)$ foregone gains by Satterthwaite and Williams \cite{satterthwaite89rate-convergence} may suggest that the coefficient of $1/m^2$ in the actual foregone gain function is small.} Moreover, parameter choices, which according to an incomplete information analysis, should greatly reduce efficiency in \acro{da}s had no such effect in recent laboratory tests \cite{friedman93@double-auction-institution-survey}.

%\clearpage

%%%%%%%%%%%%%%%%%%%%%%%%%%%%%%%%%%%%%%%%%%%%%%%%%%%%%%%%%%%%%%%%%%%%%%
\section{Experimental approaches}
\label{sec:experimental}
%%%%%%%%%%%%%%%%%%%%%%%%%%%%%%%%%%%%%%%%%%%%%%%%%%%%%%%%%%%%%%%%%%%%%%

Due to the difficulty of applying game-theoretic methods to complex auction mechanisms, researchers from economics and computer science have turned to running laboratory experiments, studying the dynamics of price formation and how the surprisingly high efficiency is obtained in a \acro{da} where information is scattered between the traders.

%---------------------------------------------------------------------
\subsection{Different data sources}
\label{sec:experimental:data-sources}
%---------------------------------------------------------------------

The data on which researchers base their studies may come from three sources: (1) field data from large-scale on-going markets, (2) laboratory data from small-scale auctions with human subjects, and (3) computer simulation experiments.

Field data has the most relevance to the real-world economy, but does not reveal many important values, e.g., the private values of traders, and hence puts limits on what can be done. The human subjects in laboratory experiments presumably inherit the same level of intelligence and incentive to make profit as in real markets, and the experiments are run under the rules that researchers aim to study. Such experiments, however, are expensive in terms of time\footnote{The experiments are run using a physical clock and need take into consideration the response time of human traders.} and money\footnote{Usually human subjects are monetarily rewarded according to their performance.} needed.

Computer-aided simulation is a less expensive alternative and can be repeated as many times as needed. However traders' strategies are not endogenously chosen as in auctions with human traders, but are specified exogenously by the experiment designers, which raises the question of whether the conclusions of this approach are trustworthy and applicable to practical situations. Gode and Sunder \cite{gode93@zi} invented a zero-intelligence strategy (or \acro{zi}) that always randomly picks a profitable price to bid or ask. Surprisingly, their experiments with \acro{cda}s exhibit high efficiency despite the lack of intelligence of the traders. Thereafter, much more work followed this path, and has gained tremendous momentum, especially given that real-world stock exchanges are becoming automated, e-business becomes an everyday activity, and the Internet reaches every corner of the globe.

%
%%---------------------------------------------------------------------
%\subsection{Typical auction simulation configuration}
%\label{sec:experimental:configuration}
%%---------------------------------------------------------------------
%
%\todo

%---------------------------------------------------------------------
\subsection{Smith's experiments}
\label{sec:experimental:smith}
%---------------------------------------------------------------------

Smith pioneered the research falling into the so-called \emph{experimental economics} field by running a series of experiments with human subjects \cite{smith62@competitive-market-behavior}. The experimental results revealed many of the properties of \acro{cda}s, which have been the basis and benchmark for much subsequent work. Smith showed that in many different cases even a handful of traders can lead to high allocative efficiency, and transaction prices can quickly converge to the theoretical equilibrium.

Smith's experiments are set up as follows:
\begin{itemize}
\item Every trader, either a buyer or a seller, is given a private value. The set of private values form the supply and demand curves.
\item Each experiment was run over a sequence of trading days, or periods,\footnote{Smith used the term \emph{periods} to refer to what is called \emph{days} in this report.} the length of which depend on how many traders are involved but are typically several minutes in duration. Different experiments may have different numbers of periods.
\item For simplicity, in most experiments, a trader is allowed to make a transaction for the exchange of only a single commodity in each day.
\item Traders are free at any time to make a bid/ask or to accept a bid/ask.
\item Once a transaction occurs, the transaction price, as well as the two traders' private values, are recorded.
\item For each new day, a trader may make up to one transaction with the same private value as before no matter whether she has made one in the previous day. Thus the supply and demand curves each correspond to a trading day. The experimental conditions of supply and demand are held constant over several successive trading days in order to give any equilibrating mechanisms an opportunity to establish an equilibrium over time, unless it is the aim to study the effect of changing conditions on market behavior.
\end{itemize}

\begin{figure*}[!tb]
  \begin{center}
  \mbox{
    \subfigure{\includegraphics[scale=0.6]{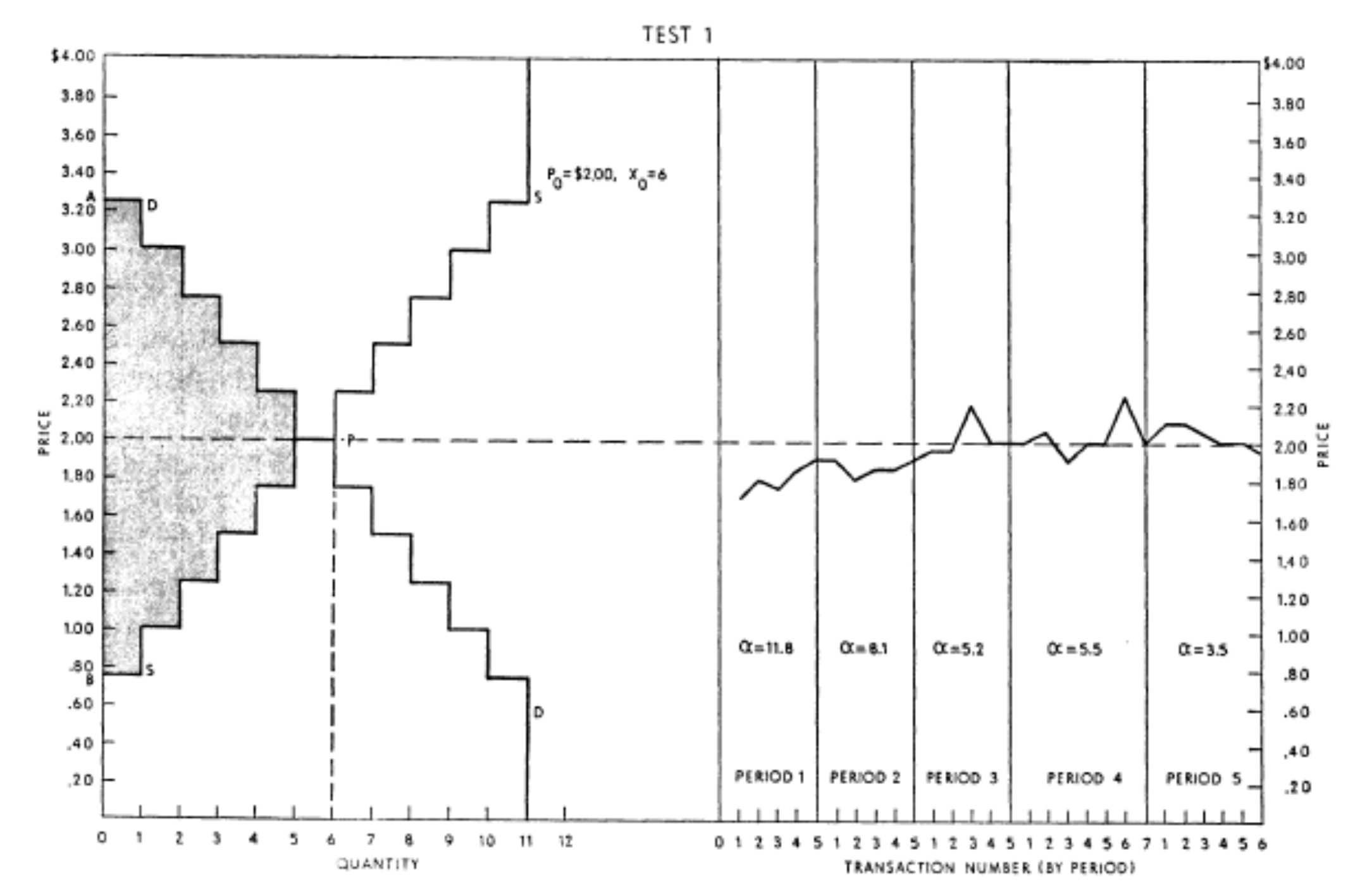}}
  }
  \caption{Supply and demand curves (left) and transaction price trajectory (right) in Smith's test 1. Originally as Chart 1 in \cite{smith62@competitive-market-behavior}. }
  \label{fig:smith-test1}
  \end{center}
\end{figure*}

\cite{smith62@competitive-market-behavior} reports 10 experiments that we discuss below. Each experiment was summarized by a diagram showing the series of transactions in the order in which they occurred. Figure~\ref{fig:smith-test1} gives one of Smith's diagrams.\footnote{Supply and demand curves of a market are typically stepped due to the discrete numbers of commodities, but the ones in Figure~\ref{fig:supply-demand} are straight line segments because it is assumed there that a large number of traders participate in the auction and thus the step-changes can be treated as infinitesimal.}. In the right-hand part of the diagram, each tick represents a transaction, rather than a unit of physical time.

Trading prices in most experiments have a striking tendency to converge on the theoretical prices, marked with a dashed line in Figure~\ref{fig:smith-test1}. To measure the tendency to converge, Smith introduced the coefficient of convergence, $\alpha$, from (\ref{equ:alpha}). Figure~\ref{fig:smith-test1} shows $\alpha$ tends to decline from one trading day to the next.

The equilibrium price and quantity of experiments 2 and 3 are approximately the same, but the latter, with the steeper inclination of supply and demand curves, converges more slowly. This complies with the Walrasian hypothesis that the rate of increase in exchange price is an increasing function of the excess demand at that price.

Experiment 4 presents an extreme case with a flat supply curve, whose result also confirms the Walrasian hypothesis, but it converges to a fairly stable price above the predicted equilibrium. In this experiment, a decrease in demand is ineffective in shocking the market down to the equilibrium. The result shows that the equilibrium may depend not only on the intersection of the supply and demand schedules, but also  upon the shapes of the schedules.

A hypothesis aiming to explain the phenomenon is that the actual market equilibrium will be above the equilibrium by an amount which depends upon how large the \emph{buyers' rent}\footnote{The area enclosed by the horizontal line at $P_0$, price axis, and the demand curve.} is relative to the \emph{sellers' rent}.\footnote{The area enclosed by the horizontal line at $P_0$, price axis, and the supply curve.} Experiment 7, which was designed with the purpose of supporting or contradicting this hypothesis, shows slow convergence, complying with the Walrasian hypothesis, but still exhibits a gradual approach to equilibrium. It is concluded that a still smaller buyers' rent may be required to provide any clear downward bias in the static equilibrium. What's more, it seems ``quite unmistakable" that the bigger the difference between the buyers' rent and sellers' rent, the slower the convergence. Smith speculated that the lack of monetary payoffs to the experimental traders may have an effect on the markets. A strong measure to further test the hypothesis is to mimic real markets as exactly as possible by paying each trader a small return just for making a contract in any period, which according to some experiments induces faster convergence.

Experiment 5 was designed to study the effect on market behavior of changes in the conditions of demand and supply. At some point in the experiment, new buyers were introduced resulting in an increase in demand. The eagerness to buy causes the trading price to increase substantially once the market resumes and the price surpasses the previous equilibrium.

Experiment 6 was designed to determine whether market equilibrium was affected by a marked imbalance between the number of intra-marginal sellers and the number of intra-marginal buyers near the predicted equilibrium price. The result confirmed the effect of a divergence between buyer and seller rent on the approach to equilibrium, but the lack of marginal sellers near the theoretical equilibrium did not prevent the equilibrium from being attained. The change of decrease in demand at the end of the fourth trading day showed that the market responded promptly by showing apparent convergence to the new, lower, equilibrium.

In contrast to the previous experiments, the market in experiment 8 was designed to simulate an ordinary retail market, in which only sellers are allowed to enunciate offers, and buyers could only either accept or reject the offers of sellers. Due to the desire of sellers to sell at higher prices, the trading prices in the first period remained above the predicted equilibrium. But starting at the second period, the trading price decreased significantly and remained below the equilibrium, not only because the early buyers again refrained from accepting any high price offers, but also because the competition among sellers became more intense. Later in the experiment, when the previous market pricing organization was resumed, exchange prices immediately moved toward equilibrium.

Experiments 9 and 10 are similar to experiment 7 except that each trader is allowed to make up to 2 transactions with the assigned private value within each day. The results showed that the increase in volume helps to speed up the convergence to equilibrium. The same results were obtained even when demand was increased during experiment 9.

%\clearpage

%%%%%%%%%%%%%%%%%%%%%%%%%%%%%%%%%%%%%%%%%%%%%%%%%%%%%%%%%%%%%%%%%%%%%%
\section{Trading agents}
\label{sec:trading}
%%%%%%%%%%%%%%%%%%%%%%%%%%%%%%%%%%%%%%%%%%%%%%%%%%%%%%%%%%%%%%%%%%%%%%

%---------------------------------------------------------------------
\subsection{Zero intelligence traders}
\label{sec:trading:zi}
%---------------------------------------------------------------------

Smith's focus in \cite{smith62@competitive-market-behavior} was mainly on the convergence of transaction prices in different scenarios rather than directly examining why high efficiency is obtained. However, high efficiency is usually the goal of a \acro{da} market designer. In a computerized world, a question that arises naturally is whether Smith's results can be replicated in electronic auctions. In Smith's experiments, as is traditional in real markets, the traders are human beings, but computer programs are supposed to be automatic and work without human involvement. Obviously humans are intelligent creatures, but programs are not, at least for the foreseeable future. Is it intelligence that contributes to the high efficiency of double auction markets, or is it something else?

Gode and Sunder \cite{gode93@zi, gode93@lower-bound} were among the first to address this question, claiming that no intelligence is necessary for the goal of achieving high efficiency; so the outcome is due to the auction mechanism itself.

They reached this position having introduced two trading strategies: \emph{zero intelligence without constraint} or \textsc{zi-u} and \emph{zero intelligence with constraint} or \textsc{zi-c}. \textsc{zi-u}, the more na\"ive version, shouts an offer at a random price without considering whether it is losing money or not, while \textsc{zi-c}, which lacks the motivation of maximizing profit and picks a price in a similar way to \textsc{zi-u}, simply makes shouts that guarantee no loss.

It was shown that \acro{zi-u} performs poorly in terms of making a profit, but \acro{zi-c} generates high efficiency solutions, comparable to the human markets (see Table~\ref{table:zi-eff}) and can be considered to place a lower bound on the efficiency of markets \cite{gode93@lower-bound}.
\begin{table}[!b]
  \centering
  \begin{tabular}{lccccc}
    \hline \hline
    Traders & Market 1 & Market 2 & Market 3 & Market 4 & Market 5 \\ \hline
    \acro{zi-u} & 90.0 & 90.0 & 76.7 & 48.8 & 86.0 \\
    \acro{zi-c} & 99.9 & 99.2 & 99.0 & 98.2 & 97.1 \\
    Human & 99.7 & 99.1 & 100.0 & 99.1 & 90.2 \\
    \hline
  \end{tabular}
  \caption{Mean efficiency of markets in Gode and Sunder's experiments. Originally as Table 2 in
\cite{gode93@zi}.}\label{table:zi-eff}
\end{table}

\begin{figure*}[!tb]
  \begin{center}
  \mbox{
    \subfigure{\includegraphics[scale=0.6]{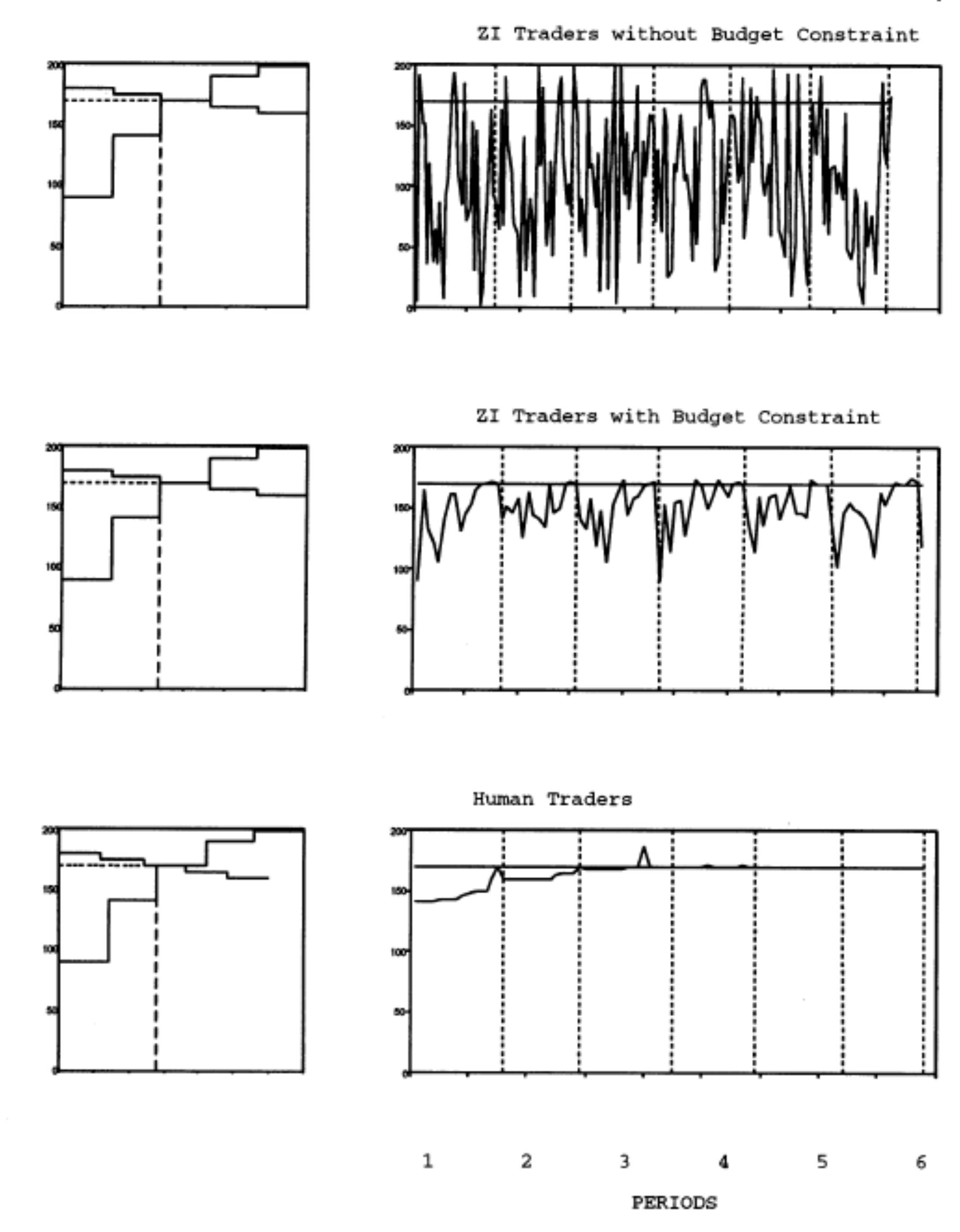}}
  }
  \caption{Gode and Sunder's experiments comparing \acro{zi-u} traders (top), \acro{zi-c} traders (middle), and human traders (bottom). Originally as Fig. 4 in \cite{gode93@zi}.}
  \label{fig:zi-test4}
  \end{center}
\end{figure*}

Gode and Sunder's experiments were setup with similar rules as in Smith's. They designed five different supply and demand schedules and tested each of them respectively with the three kinds of homogeneous traders, \acro{zi-u}, \acro{zi-c}, and human traders. Figure~\ref{fig:zi-test4} presents what happened in one of their experiments.

Prices in the \acro{zi-u} market exhibit little systematic pattern and no tendency to converge toward any specific level, but on the contrary, prices in the human market, after some initial adjustments, settle in the proximity of the equilibrium price (indicated by a solid horizontal line in all panels in Figure~\ref{fig:zi-test4}). Gode and Sunder then raised the question: how much of the difference between the market outcomes with \acro{zi-u} traders and those with human traders is attributable to intelligence and profit motivation, and how much is attributable to market discipline?

They argue that, after examining the performance of the \acro{zi-c} markets, it is market discipline that plays a major role in achieving high efficiency. Though in the \acro{zi-c} market, the price series shows no signs of improving from day to day, and the volatility of the price series is greater than the volatility of the price series from the human market, the series converges slowly toward equilibrium within each day. Gode and Suner's explanation is that it is due to the progressive narrowing of the opportunity sets of \acro{zi-c} traders, e.g., the set of intra-marginal traders. Despite the randomness of \acro{zi-c}, buyers with higher private values tend to generate higher offered prices and they are likely to trade with sellers earlier than those buyers further down the demand curve. A similar statement also holds for sellers. Thus as the auction goes on, the upper end of the demand curve shifts down and the lower end of the supply curve moves up, which means the feasible range of transaction prices narrows as more commodities are traded, and transaction prices will converge to the equilibrium price. The fact that \acro{zi-c} traders lack profit motivation and have only the minimal intelligence (just enough to avoid losing money) suggests that the market mechanism is the key to obtaining high efficiency.

%\clearpage

%---------------------------------------------------------------------
\subsection{Zero intelligence plus and beyond}
\label{sec:trading:zip-misc}
%---------------------------------------------------------------------

Gode and Sunder's results were, however, questioned by Cliff and Bruten \cite{cliff97@minimal-intell-agent}. The latter agreed on the point that the market mechanism plays a major role in achieving high efficiency, but disputed whether in \acro{zi-c} markets transaction prices will always converge on equilibrium price. They argued that the mean or expected value of the transaction price distribution was shown quantitatively to get close to the equilibrium price only in situations where the magnitude of the gradient of linear supply and demand curves is roughly equal, and used this to infer that zero-intelligence traders are not sufficient to account for convergence to equilibrium.

Cliff and Bruten further designed an adaptive trading strategy called \emph{zero intelligence plus} or \textsc{zip}. Like \acro{zi-c}, \acro{zip} traders make stochastic bids, but can adjust their prices based on the auction history, i.e., rasing or lowering their profit margins dynamically according to the actions of other traders in the market. More specifically, \acro{zip} traders raise the profit margin when a less competitive offer from the competition\footnote{That is sellers compete against sellers to get asks accepted and buyers compete against buyers to get bids accepted.} is accepted, and lower the profit margin when a more competitive offer from the competition is rejected, or an accepted offer from the other side of the market would have been rejected by the subject. At every step, the profit margin is updated according to a learning algorithm called the \emph{Widrow-Hoff delta rule} in which a value being learned is adapted gradually towards a moving target, and the past targets leave discounting momentum to some extent.

Cliff and Bruten concluded that the performance of \textsc{zip} traders in the experimental markets is significantly closer to that of human traders than is the performance of \acro{zi-c} traders, based on the observation that \acro{zip} traders rapidly adapt to give profit dispersion\footnote{Profit dispersion is the root mean squared difference between actual and equilibrium profits, and can be expressed as $$\sqrt{\frac{1}{n}\Sigma_i{(a_i-\pi_i)^2}},$$ where $a_i$ and $\pi_i$ are the actual and theoretical equilibrium profits of trader $i$, $i=1,\cdots,n$.} levels that are in some cases approximately a factor of ten less than those of \acro{zi-c} traders.

Preist and van Tol introduced a revised version of \acro{zip}, which we call \acro{pvt}, and reported faster convergence to equilibrium and robustness to changes in parameter configuration \cite{preist98@adaptive-agent-in-persistent-shout-da}.

Other learning methods have been adopted to design even more complex trading strategies than \acro{zip} and its variants. Roth and Erev \cite{roth95learning} proposed a reinforcement-based stimuli-response strategy, which we call \textsc{re}. \acro{re} traders adapt their trading behavior in successive auction rounds by using their profits in the last round as a reward signal. Gjerstad and Dickhaut \cite{gjerstad98@gd} suggested a best-response-based strategy, which is commonly referred to as \textsc{gd}. \acro{gd} traders keep a sliding window of the history of the shouts and transactions and calculate the probabilities of their offers being accepted at different prices. The traders use a cubic interpolation on the shouts and transaction prices in the sliding window in order to compute the probability of future shouts being accepted. They then use this to calculate the expected profit of those shouts. The expected profit at a price is the product of the probability of the price being accepted and the difference between the price and the private value. \acro{gd} traders then always choose to bid or ask at a price that maximizes their expected profit. \acro{gd} is the most computation-intensive trading strategy considered so far, and indeed generates the best record both for allocative efficiency and the speed of convergence to equilibrium compared to the other trading strategies in literature.

\begin{figure*}[!tb]
  \begin{center}
  \mbox{
    \subfigure{\includegraphics[scale=0.5]{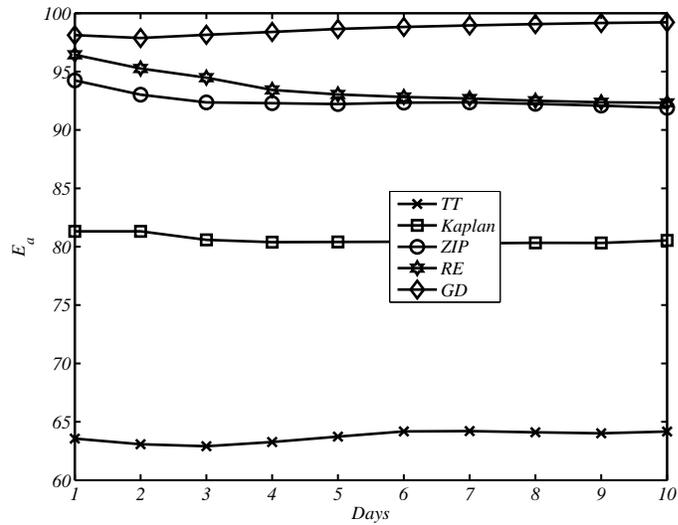}}
  }
  \caption{$E_a$s for \acro{cda} markets populated by common trading strategies}
  \label{fig:eff}
  \end{center}
\end{figure*}

\begin{figure*}[!tb]
  \begin{center}
  \mbox{
    \subfigure{\includegraphics[scale=0.5]{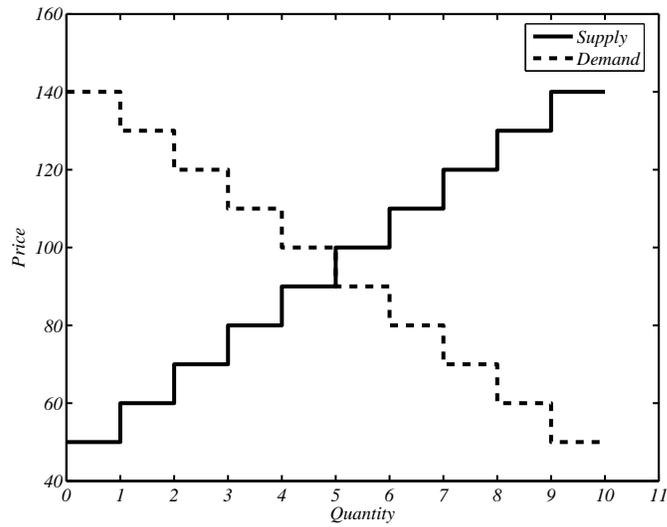}}
  }
  \caption{Supply and demand schedules in the \acro{cda} markets presented in Figure~\ref{fig:eff}.}
  \label{fig:ds}
  \end{center}
\end{figure*}

By way of indicating typical efficiencies achieved in a \textsc{cda}, Figure~\ref{fig:eff} shows the trend of the overall efficiencies of homogeneous \textsc{cda}s lasting 10 days with 50 rounds per day in which 10 buyers and 10 sellers all use the same strategy, one of: \textsc{tt},\footnote{\textsc{tt} denotes the Truth-telling strategy, in which agents truthfully report their private values.} \acro{Kaplan},\footnote{`Kaplan' refers to Todd Kaplan's sniping strategy, in which agents wait until the last minute before attempting to steal the deal \cite{rust92@trading-automata}. See Section~\ref{sec:trading:interaction-heterogeneous} for more information.} \textsc{zip}, \textsc{re}, and \textsc{gd}. The results are averaged over 400 iterations and obtained in \textsc{jasa}---the extensible Java-based auction simulation environment \cite{jasa}. Figure~\ref{fig:ds} gives the supply and demand schedules in the markets.

%---------------------------------------------------------------------
\subsection{Interaction of heterogenous trading strategies}
\label{sec:trading:interaction-heterogeneous}
%---------------------------------------------------------------------

All the above empirical works have employed either human traders or homogeneous trading agents, demonstrating high efficiency and fast convergence to equilibrium, and some of this work has also produced theoretical results. It is however necessary to see how an auction works populated by heterogeneous trading subjects.

There are both theoretical and practical reasons for considering heterogeneous traders. As Rust \emph{et al.} argued in \cite{rust92@trading-automata}:
\begin{quotation}
Although current theories of \acro{da} markets have provided important insight into the nature of trading strategies and price formation, it is fair to say that none of them has provided a satisfactory resolution of ``Hayek's problem".\footnote{That is how the trading process aggregates traders' dispersed information, driving the market towards competitive equilibrium.} In particular, current theories assume a substantial degree of implicit coordination by requiring that traders have common knowledge of each other's strategies (in game-theoretic models), or by assuming that all traders use the same strategy (in learning models). Little is known theoretically about price formation in \acro{da} markets populated by heterogeneous traders with limited knowledge of their opponents.

... the assumption that players have common knowledge of each other's beliefs and strategies ... presumes an unreasonably high degree of implicit coordination amongst the traders ... Game theory also assumes that there is no \textit{a priori} bound on traders' ability to compute their \acro{bne} strategies. However, even traders with infinite, costless computing capabilities may still decide to deviate from their \acro{bne} strategies if they believe that limitations of other traders force them to use a sub-optimal strategy.
\end{quotation}

They went on to argue that \acro{zi-c} and other strategies' striking performance strongly suggests that the nice properties have more to do with the market mechanism itself than the rationality of traders. In addition, strategies that are more individually rational than \acro{zi-c} may display less collective rationality since clever strategies can exploit unsophisticated ones such as \acro{tt} and \acro{zi-c} so that a more-intelligent extra-marginal trader has more chances to finagle a transaction with an intra-marginal traders, causing market efficiency to fall.

To observe heterogeneous auctions, the Santa Fe Double Auction Tournament (\acro{sfdat}) was held in 1990 and prizes were offered to entrants in proportion to the trading profits earned by their programs over the course of the tournament. 30 programs from researchers in various fields and industry participated. The majority of the programs encoded the entrant's ``market intuition" using simple rules of thumb. The top-ranked program was \acro{kaplan}, named after the entrant. \acro{kaplan} and the runner-up strategy are remarkably similar. Both ``wait in the background and let the others do the negotiating, but when bid and ask get sufficiently close, jump in and steal the deal" \cite{rust92@trading-automata}.

The overall efficiency levels in the markets used in the tournaments originally appear to be somewhat lower than that observed in experimental markets with human traders, but experiments without the last-placed players produced an efficiency of around 97\%. This is further evidence that the properties of traders also affect the outcome of \acro{da} markets to some extent.

Besides high efficiency levels and convergence to competitive equilibrium, other ``stylized facts" of human \acro{da} markets observed in the \acro{sfdat} include: reductions in transaction-price volatility and efficiency losses in successive trading days that seem to reflect apparent learning effects, coexistence of extra-marginal and intra-marginal efficiency losses, and low-rank correlations between the \emph{realized order of transactions} and the \emph{efficient order}.\footnote{The \emph{efficient order} is the transaction sequence that maximizes surplus, meaning that the first transaction occurs between the buyer with the highest private value and the seller with the lowest private value, the second transaction occurs between the buyer and seller next to them, and so on. The \emph{realized order of transactions} is the actual order in which transactions are made.}

Thorough examination of efficiency losses in the tournaments and later experiments indicates that the success of \acro{kaplan} is due to its patience in waiting to exploit the intelligence or stupidity of other trading strategies.\footnote{The usual higher efficiency of \acro{ch}s than \acro{cda}s can also be viewed as the proactive elimination of the effect of traders' impatience.}

The volume of e-commerce nowadays creates another motivation for evaluating trading strategies in a heterogeneous environment. Electronic agents, on behalf of their human owners, can automatically make strategic decisions and respond quickly to the changes in various kinds of markets. In the foreseeable future, these agents will have to compete with a variety of agents using a range of trading strategies and human traders. As more complex trading strategies appear, it is natural to speculate on how these electronic minds will compete against their human counterparts.

Das \textit{et al.} ran a series of \acro{cda}s allowing persistent orders\footnote{In the \acro{sfdat} and the \acro{cda} testing \acro{zip} in \cite{cliff97@minimal-intell-agent}, shouts that are outbid are removed from the market, which is however not typical of real marketplaces.} populated by a mixed population of automated agents (using modified \acro{gd} and \acro{zip} strategies) and human traders \cite{das01@agent-human-interaction-in-cda}. They found that though the efficiency of the \acro{cda}s was comparable with prior research, the agents outperformed the humans in all the experiments, obtaining about 20\% more profit. Das \textit{et al.} speculated that this was due to human errors or weakness, and human traders were observed to improve their performance as they got familiar with using the trading software. Das \textit{et al.} also suggested that the weaknesses of trading agents may be found when human experts take them on and thus improvement can be made to the algorithms of the trading agents.\footnote{\cite{das01@agent-human-interaction-in-cda} also reported that either buyers consistently exploited sellers, or vice versa. However no convincing analysis was given. Similar phenomenon also occurred in experiments described in \cite{phelps03optimizing.pricing.rules}. It is not clear whether this is caused by the inherent randomness in the trading agents.}

Tesauro and Das \cite{tesauro01@high-performance-bidding-agents-4-CDA} executed experiments with both homogeneous and heterogeneous trading agents with varying trader population composition, making it possible to gain more insights into the relative competitiveness of trading strategies. In either the so-called ``one-in-many"\footnote{A single agent of one type competes against an otherwise homogeneous population of a different type.} tests or ``balanced-group"\footnote{Buyers and sellers are evenly split between two types, and every agent of one type has a counterpart of the other type with identical limit prices.} tests, \acro{gd} and \acro{zip} (and their variants) exhibited superior performance over \acro{zi-c} and \acro{kaplan} even when the market mechanisms vary to some extent.\footnote{\cite{tesauro01@high-performance-bidding-agents-4-CDA} tested both with and without the \acro{nyse} shout improvement rule, a standing shout queue, and allowance of shout modification.} Furthermore, \acro{mgd}, a variant of \acro{gd} due to Das \textit{et al.} \cite{das01@agent-human-interaction-in-cda}, outperformed all the other strategies.

The above approaches nevertheless all employ a fixed competition environment. In practice, when a strategy dominates others, it tends to flourish and be adopted by more people. Rust \textit{et al.} are the first that we are aware of to conduct evolutionary experiments, where the relative numbers of the different trading strategies changed over time, so that more profitable strategies became more numerous than less profitable ones. Such an analysis revealed that although \acro{kaplan} agents outperformed others when traders of different types are approximately evenly distributed, they later exhibited low overall efficiency as they became the majority, making the evolution process a cycle of ups and downs.

\begin{figure*}[!tb]
  \begin{center}
  \mbox{
    \subfigure[before
perturbation]{\label{fig:walsh-before-perturbation}\includegraphics[scale=0.45]{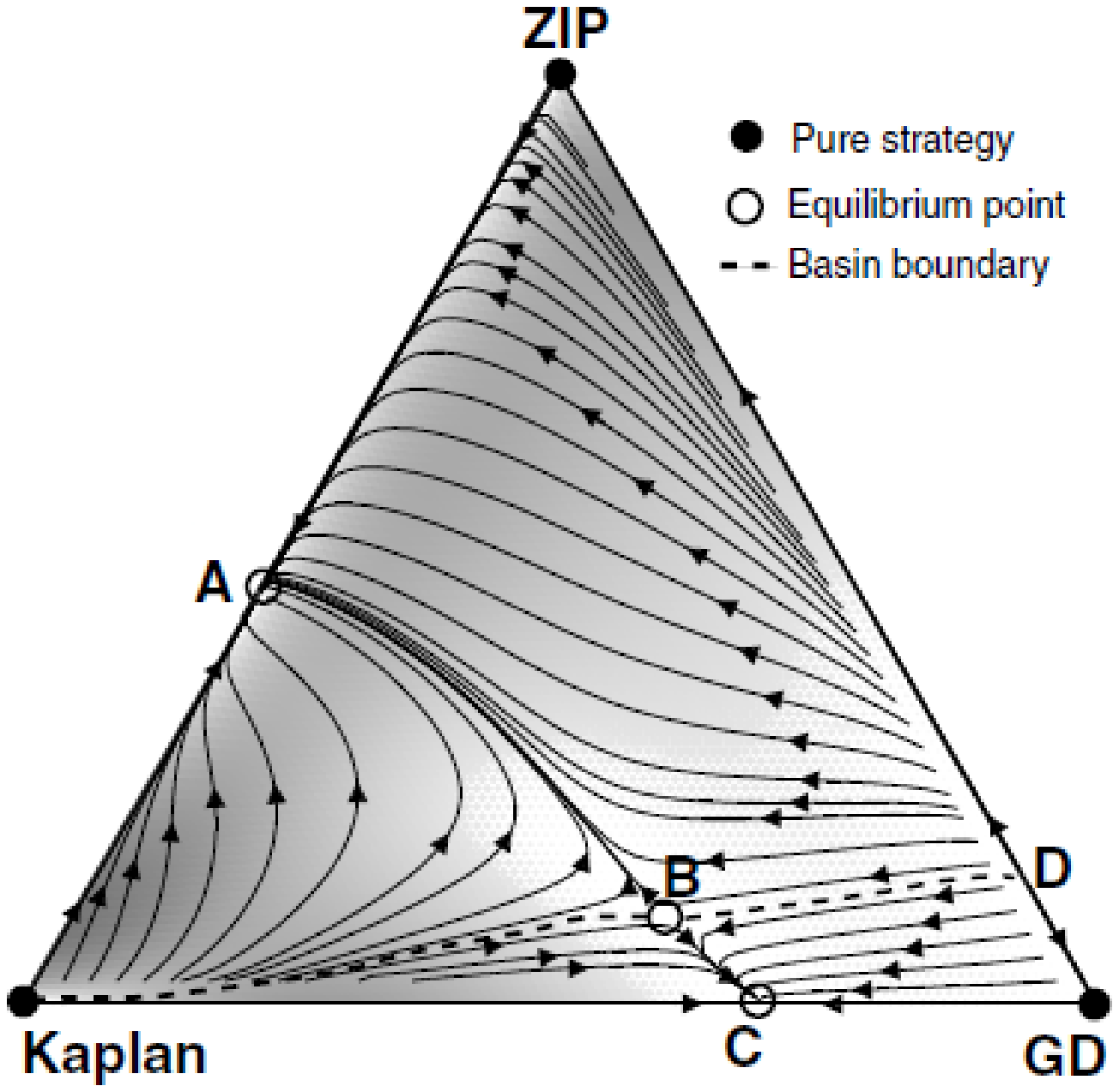}}
    \subfigure[after
perturbation]{\label{fig:walsh-after-perturbation}\includegraphics[scale=0.45]{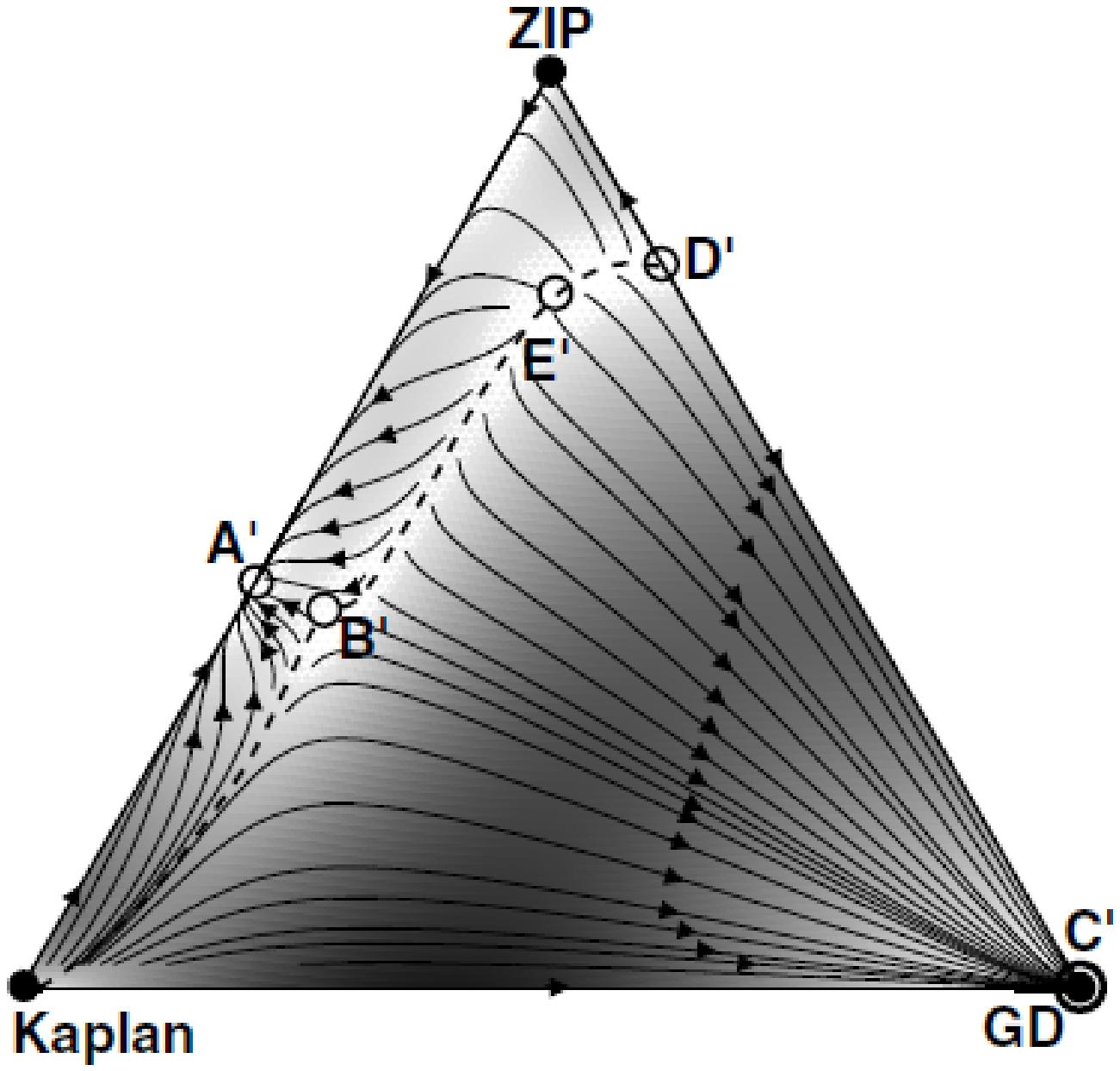}}
  }
  \caption{The replicator dynamics of \acro{cda} with \acro{zip}, \acro{kaplan}, and \acro{gd}. Originally as Figure 2
in
\cite{walsh02@analyzing-complex-strategic-interactions}.}
  \label{fig:walsh-replicator-dynamics}
  \end{center}
\end{figure*}

Walsh \textit{et al.} \cite{walsh02@analyzing-complex-strategic-interactions} gave a more formal analysis combining the game-theoretic solution concept of \acro{ne} and \emph{replicator dynamics}. They treated heuristic strategies, rather than the atomic actions like a bid or ask, as primitive, and computed expected payoffs of each individual strategy at certain points of the joint heuristic strategy space.\footnote{That is a space of a mixture of strategies when their relative proportions vary.} This method reduced the model of the game from a potentially very complex, multi-stage game to a one-shot game in normal form. At points where one strategy gains more than others, replicator dynamics dictates that the whole population moves to a nearby point where the winning strategy takes a larger fraction of the population. This process continues until an equilibrium point is reached where either the population becomes homogeneous or all strategies are equally competitive in terms of their expected payoffs. There may be multiple equilibrium points\footnote{Each equilibrium point also represents a mixed strategy, a homogeneous population of which makes a \acro{ne}.} `absorbing' areas of different sizes, \emph{basins} of the equilibria, which together compose the whole strategy space. In particular, Figure~\ref{fig:walsh-before-perturbation} shows the replicator dynamics of a \acro{cda} market with three strategies. $A$, $B$, $C$, and $D$ are all equilibrium points, but $B$ and $D$ are not stable since a small deviation from them will lead to one of the other equilibria. The triangle field gives an overview of the interaction of the three strategies and their relative competitiveness. What's more, a technique called \emph{perturbation analysis} is used to evaluate the potential to improve on a strategy. Figure~\ref{fig:walsh-after-perturbation} shows the replicator dynamics of the same strategies after small portions of both \acro{zip} and \acro{kaplan}'s payoffs were shifted to \acro{gd}. Such a shift significantly changed the landscape of the space, and \acro{gd} dominated in most of possible combinations. This showed that a `tiny' improvement on the \acro{gd} strategy may greatly affect its competition against the other strategies.

Phelps \textit{et al.} \cite{phelps06automatic-strategy-acquisition, phelps05@better-response-strategies-for-da} took a similar approach in comparing the \acro{re}, \acro{tt}, and \acro{gd} strategies, showed the potential of \acro{re}, and demonstrated that a modified \acro{re} strategy could be evolved by optimizing its learning component.

The main drawback of this approach is an exponential dependence on the number of strategies, which limits its applicability to real-world domains where there are potentially many heuristic strategies. Walsh \textit{et al.} \cite{walsh03heuristic-strategy-ne} proposed information theoretic approaches to deliberately choose the sample points in the strategy space through an interleaving of equilibrium calculations and payoff refinement, thus reducing the number of samples required.

%---------------------------------------------------------------------
\subsection{Automating strategy acquisition}
\label{sec:trading:automating}
%---------------------------------------------------------------------

Designing heuristic strategies to a great extent depends on the intelligence and experience of the strategy designer. Prior studies have also demonstrated that heuristic strategies' performance hinges on the selection of parameter values. Automatic optimization is preferable in this sense to find best parameter combinations and further identify better strategies. Cliff and Phelps \textit{et al.} are the pioneers in this work, adopting evolutionary computation to address the challenge.

%---------------------------------------------------------------------
\subsubsection{Evolutionary computation}
\label{sec:trading:automating:ec}
%---------------------------------------------------------------------

A \emph{genetic algorithm} (or \acro{ga}) is a search technique used in computing to find true or approximate solutions to optimization and search problems. Genetic algorithms are a particular class of evolutionary algorithms that use techniques inspired by evolutionary biology such as \emph{inheritance}, \emph{mutation}, \emph{selection}, and \emph{crossover} \cite{wikipedia-site}.

A typical genetic algorithm requires two things to be defined:
\begin{enumerate}
\item a \emph{genetic representation} of the solution domain, also called the \emph{genotype} or \emph{chromosome} of the solution species,
\item a \emph{fitness function} to evaluate the solution domain.
\end{enumerate}

A standard representation of the solution is as an array of bits. Arrays of other types and structures can be used in essentially the same way. The main property that makes these genetic representations convenient is that their parts are easily aligned due to their fixed size, which facilitates simple crossover operation. Variable length representations have also been used, but crossover implementation is more complex in this case.

The fitness function is defined over the genetic representation of a solution and measures the quality of the solution. The fitness function is always problem dependent. For instance, in the knapsack problem we want to maximize the total value of objects that we can put in a knapsack of some fixed capacity. A representation of a solution might be an array of bits, where each bit represents a different object, and the value of the bit (0 or 1) represents whether or not the object is in the knapsack. Not every such representation is valid, as the size of objects may exceed the capacity of the knapsack. The fitness of the solution is the sum of values of all objects in the knapsack if the representation is valid, or 0 otherwise. In some problems, it is hard or even impossible to define the fitness expression; in these cases, interactive genetic algorithms are used.

Once we have the genetic representation and the fitness function defined, the \acro{ga} proceeds to initialize a population of solutions randomly, then improve it through repetitive application of mutation, crossover, and selection operators.

%---------------------------------------------------------------------
\subsubsection{Optimizing parameter combination in \acro{zip}}
\label{sec:trading:automating:cliff}
%---------------------------------------------------------------------

Cliff addressed the labor-intensive manual parameter optimization for the \acro{zip} strategy, automatically optimizing parameter selection using a \acro{ga} \cite{cliff01@ev-op-param-for-agents}. He identified eight parameters in \acro{zip}: lower and upper bounds of the learning rate $\beta$ (how fast to move towards the target), momentum $\gamma$ (how much past momentum to carry over), and initial profit margin $\mu$, and the upper bounds of the ranges defining the distributions of absolute and relative perturbations on learned prices, respectively denoted as $c_a$ and $c_r$. These real parameters make an eight-dimensional space and any parameter value combination corresponds to a point in that space. The vector of the eight parameters defines an ideal genotype.

%---------------------------------------------------------------------
\subsubsection{Combining \acro{ga} and heuristic strategy analysis}
\label{sec:trading:automating:phelps}
%---------------------------------------------------------------------

Phelps \textit{et al.} took a step further along this track. They combined the replicator-dynamics-based heuristic strategy analysis method in \cite{walsh02@analyzing-complex-strategic-interactions} and a \acro{ga}, identified a strategy as the basis for optimization, and successfully evolved the strategy and acquired an optimized strategy that can beat \acro{gd}, commonly considered the most competitive strategy \cite{phelps06automatic-strategy-acquisition, phelps05@better-response-strategies-for-da}.

Since it is not realistic to seek ``best", or even ``good", strategies that can beat all potential opponents because an absolutely dominating strategy does not appear to exist in the \acro{cda} trading scenario---since the performance of a strategy depends greatly on the types of the opponents---Phelps \textit{et al.} proposed using a small finite population of randomly sampled strategies to approximate the game with an infinite strategy population consisting of a mixture of all possible strategies. In particular, \acro{re}, \acro{tt}, and \acro{gd} were chosen as sample strategies. Following the heuristic strategy analysis and perturbation method in \cite{walsh02@analyzing-complex-strategic-interactions}, \acro{re} was found to have the potential to dominate \acro{tt} and \acro{gd}.

The \acro{re} strategy uses reinforcement learning to choose from $n$ possible profit margins over the agent's private value based on a reward signal computed as a function of profits earned in the previous round of bidding. Potentially, the \acro{re} learning algorithm may be replaced by a number of learning algorithms, including \acro{sq} (stateless Q-learning), \acro{npt} (a modified version of \acro{re} used in \cite{nicolaisen01market-power}), and \acro{dr} (a control algorithm which selects a uniformly random action regardless of reward signal). Phelps \textit{et al.} then encoded the genotype to select any of these algorithms together with their parameters. The evolutionary search procedure they used is similar to Cliff's except that the individuals in a generation are evaluated again with the heuristic strategy analysis approach and the basin size is used as a measure of fitness. The experiment finally found a \acro{sq} algorithm with a particular parameter combination, which together with \acro{tt} composes the Nash equilibrium that captures 97\% of the strategy space populated by the learned strategy, \acro{tt}, \acro{re}, and \acro{gd}.

%---------------------------------------------------------------------
\subsection{Trading Agent Competition}
\label{sec:trading:tac}
%---------------------------------------------------------------------

The Trading Agent Competition (\acro{tac}) was organized to promote and encourage high quality research into trading agents. Under the \acro{tac} umbrella, a series of competitions have been held, including two types of game, \acro{tac} \acro{c}lassic and \acro{tac scm} \cite{wellman03tac}.

\acro{tac} \acro{c}lassic sets up a ``travel agent" scenario based on complex procurement in multiple simultaneous auctions. Each travel agent (an entrant to the competition) has the goal of assembling travel packages (from \acro{TAC}town to Tampa, during a notional multi-day period). Each agent is acting on behalf of a certain number of clients, who express their preferences for various aspects of the trip. The objective of the travel agent is to maximize the total satisfaction of its clients (the sum of the client utilities).

\acro{tac scm} was designed to capture many of the challenges involved in supporting dynamic supply chain practices in the industry of PC manufacturing. Supply chain management is concerned with planning and coordinating the activities of organizations across the supply chain, from raw material procurement to the delivery of finished goods. In today's global economy, effective supply chain management is vital to the competitiveness of manufacturing enterprizes as it directly impacts their ability to meet changing market demands in a timely and cost effective manner. In \acro{tac scm}, agents are simulations of small manufacturers, who must compete with each other for both supplies and customers, and manage inventories and production facilities.

%\clearpage

%---------------------------------------------------------------------
\section{Experimental auction mechanism design}
\label{sec:mechdes}
%---------------------------------------------------------------------

Mechanism design applied to auctions explores how to design the rules that govern auctions to obtain specific goals.

The story of trading strategies in the preceding section is only one facet of the research on auctions. Gode and Sunder's results suggest that auction mechanisms play an important role in determining the outcome of an auction, and this is further bourne out by the work of Walsh \textit{et al.} \cite{walsh02@analyzing-complex-strategic-interactions}, which also points out that results hinge on both auction design and the mix of trading strategies used.

According to classical auction theory, if an auction is \emph{strategy-proof} or \emph{incentive compatible}, traders need not bother to conceal their private values and in such auctions complex trading agents are not required. However, typical \acro{da}s are not strategy-proof. McAfee \cite{mcafee92dominant-strategy-da} has derived a form of double auction that is strategy-proof, though this strategy-proofness comes at the cost of lower efficiency.

Despite the success of analytic approaches to the relatively simple auctions presented in Section~\ref{sec:auction-theory}, the high complexity of the dynamics of some other auction types, especially \acro{da}s, makes it difficult to go further in using analytical methods \cite{madhavan92trading-mech, satterthwaite93@bayesian-theory, walsh02@analyzing-complex-strategic-interactions}.

As a result, researchers turned to empirical approaches using machine learning techniques, sometimes combined with methods from traditional game theory. Instead of trying to design optimal auction mechanisms, the computational approach looks for relatively good auctions and aims to make them better, in a noisy economic environment with traders that are not perfectly rational.

%---------------------------------------------------------------------
\subsection{A parameterized space of auctions}
\label{sec:mechdes:parameterized-space}
%---------------------------------------------------------------------

One can think of different forms of auctions as employing variations of a common set of the auction rules, forming a parameterized auction space. Wurman \textit{et al.} and others parameterized auction rules using the following classification \cite{rothkopf01intro2auctions, wurman01parametrization, wurman:wellman:walsh02specifying-auction-rules}:
\begin{itemize}
\item Bidding rules: determine the semantic content of messages, the authority to place certain types of bids, and admissibility criteria for submission and withdrawal of bids.
  \begin{itemize}
  \item How many sellers and buyers are there?
  \item Are both groups allowed to make shouts?
  \item How is a shout expressed?
  \item Does a shout have to beat the corresponding market quote if one exists?
  \end{itemize}
\item Information revelation:
  \begin{itemize}
  \item When and what market quotes are generated and announced?
  \item Are shouts visible to all traders?
  \end{itemize}
\item Clearing policy:
  \begin{itemize}
  \item When does clearing a market take place?
  \item When does a market close?
  \item How are shouts matched?
  \item How is a transaction price determined?
  \end{itemize}
\end{itemize}

The idea of parameterizing auction space not only eases the heuristic auction mechanism design, but also makes it possible to `search' for better mechanisms in an automated manner \cite{cliff02@evolving, phelps02coev.mechanism.design}.

It is not yet clear how auction design, and thus the choice of parameter values, contributes to the observed performance of auctions. Thus it is not clear how to create an auction with a particular specification. It \emph{is} possible to design simple mechanisms in a provably correct manner from a specification, as shown by Conitzer and Sandholm \cite{conitzer03automated, conitzer04algorithm}. However it is not clear that this kind of approach can be extended to mechanisms as complex as \acro{da}s. As a result, it seems that we will have to design double auction mechanisms experimentally, at least for the foreseeable future.

Of course, doing things experimentally does not solve the general problem. A typical experimental approach is to fix all but one parameter, creating a one-dimensional space, and then measure performance across a number of discrete sample points in the space, obtaining a fitness landscape that is expected to show how the factor in question correlates to a certain type of performance and how the auction can be optimized by tweaking the value of that factor \cite{phelps03optimizing.pricing.rules}. In other words, the experimental approach examines one small part of a mechanism and tries to optimize that part.\footnote{And of course there are rarely any guarantees as to the optimality of the results.} The situation is complicated when more than one factor needs to be taken into consideration --- the search space then becomes complex and multiple dimensional, and the computation required to map and search it quickly becomes prohibitive.

%---------------------------------------------------------------------
\subsection{Evolving market mechanisms}
\label{sec:mechdes:evolving}
%---------------------------------------------------------------------

Instead of manual search, some researchers have used evolutionary computation to automate mechanism design in a way that is similar to the evolutionary approach to optimizing trading strategies.

Cliff \cite{cliff01@ev-of-market-mech} explored a continuous space of auction mechanisms by varying the probability of the next shout (at any point in time) being made by a seller, denoted by $Q_s$. The continuum includes the \textsc{cda} ($Q_s=0.5$) and also two purely single-sided mechanisms that are similar to the English auction ($Q_s=0.0$) and the Dutch auction ($Q_s=1.0$). Cliff's experiments used genetic algorithms and found that a $Q_s$ that corresponds to a completely new kind of auction led to a better $\alpha$ value than that obtained for other markets using \textsc{zip} traders. Walia \textit{et al.} \cite{walia02@evolving-market-design-with-zi} and the same authors but in a different order \cite{cliff02@evolving} continued with this work, showing that the approach is also effective in markets using \textsc{zi-c} traders, and the new ``irregular" mechanisms can lead to high efficiency with a range of different supply and demand schedules as well. The visualization of fitness landscapes, using plots including 3D histograms and contours, is also noteworthy.

Byde \cite{byde03applying} took a similar approach in studying the space of auction mechanisms between the first and second-price sealed-bid auctions. The winner's payment is determined as a weighted average of the two highest bids, with the weighting determined by the auction parameter. For a given population of bidders, the revenue-maximizing parameter is approximated by considering a number of parameter choices over the allowed range, using a \acro{ga} to learn the parameters of the bidders' strategies for each choice, and observing the resulting average revenues. For different bidder populations (varying bidder counts, risk sensitivity, and correlation of signals), different auction parameter values are found to maximize revenue.

Taking another tack, Phelps \textit{et al.} explored the use of genetic programming to determine auction mechanism rules automatically.

\emph{Genetic programming} (or \acro{gp}), another form of evolutionary computation that is similar to \acro{ga}s, evolves programs (or expressions) rather than the binary strings evolved in \acro{ga}s. This makes automatic programming possible, and in theory allows even more flexibility and effectiveness in finding optimal solutions in the domain of concern. In \acro{gp}, programs are traditionally encoded as \emph{tree structures}. Every tree node has an operator function and every terminal node has an operand, making mathematical expressions easy to evolve and evaluate. With tree structures, crossover is applied on an individual by simply switching one of its nodes with another node from another individual in the population. Mutation can replace a whole node in the selected individual, or it can replace just the information of that node. Replacing a node means replacing the whole branch. This adds greater effectiveness to the crossover and mutation operators \cite{wikipedia-site}.

Phelps \textit{et al.} \cite{phelps03optimizing.pricing.rules} demonstrated how \acro{gp} can be used to find an optimal point in a space of pricing policies, where the notion of optimality is based on allocative efficiency and trader market power. In \acro{da} markets, there are two popular pricing policies: the $k$-\acro{da} pricing rule \cite{satterthwaite93@bayesian-theory} and the uniform pricing policy. The former is clearly a discriminatory policy\footnote{That is transactions are cleared at different prices depending upon the prices of the matching bid and ask.} and may be represented as:
$$
p = k\cdot p_a + (1-k)\cdot p_b
$$
where $k\in [0,1]$, and $p_a$ and $p_b$ are ask and bid prices. The latter executes all transactions at the same price, typically the middle point of the interval between the market ask and bid quotes. Searching in the space of arithmetic combinations of shout prices and market quotes including the above two rules as special cases, led to a complex expression that is virtually indistinguishable from the $k = 0.5$ version of the $k$-\acro{da} pricing rule. This shows that the middle-point transaction pricing rule not only reflects the traditional practice but also can be technically justified.

Noting that the performance of an auction mechanism always depends on the mix of traders participating in the mechanism, and both the auction mechanism and the trading strategies may adapt themselves simultaneously, Phelps \textit{et al.} \cite{phelps02coev.mechanism.design} further investigated the use of co-evolution in optimizing auction mechanisms. They first co-evolved buyer and seller strategies and then together with auction mechanisms. The approach was able to produce outcomes with reasonable efficiency in both cases.

%---------------------------------------------------------------------
\subsection{Evaluating market mechanisms}
\label{sec:mechdes:evalutation}
%---------------------------------------------------------------------

Phelps \textit{et al.} proposed a novel way to evaluate and compare the performances of market mechanisms using heuristic strategy analysis \cite{phelps04ch.vs.cda}.

Despite the fact that the performance of an auction mechanism may vary significantly when the mechanism engages different sets of trading agents, previous research on auctions analyzed the properties of \acro{da} markets using an arbitrary selection of homogeneous trading strategies. A more sound approach is to find the equilibria of the game between the participating trading strategies and measure the auction mechanism at those equilibrium points. As Sections~\ref{sec:trading:interaction-heterogeneous} and \ref{sec:trading:automating:phelps} have discussed, the heuristic strategy analysis calculates equilibria among a representative collection of strategies. This makes the method ideal for measuring market mechanisms at those relatively stable equilibria.

The representative strategies selected by Phelps \textit{et al.} included \acro{re}, \acro{pvt}, and \acro{tt}. The replicator dynamics analysis revealed that: (1) neither the \acro{cda} nor the \acro{ch} mechanism is strategy-proof since \acro{tt} is not dominant in either market; (2) increasing the number of agents in the \acro{ch} led to the appearance of an equilibrium basin for an equilibrium near \acro{tt}, which agreed with the conclusion drawn through the approximate analysis in \cite{satterthwaite89rate-convergence} discussed in Section~\ref{sec:auction-theory:da}; and (3) the \acro{ch} has higher efficiency than the \acro{cda} in the sense that the three equilibrium points\footnote{Each falls onto one of the three pure strategies, though the sizes of their basins vary.} in the dynamics field for the \acro{ch} all generate 100\% efficiency while the only equilibrium\footnote{Pure \acro{re} strategy.} for \acro{cda} produces 98\% efficiency. One can interpret the small efficiency difference as justifying the \acro{nyse}'s use of a \acro{cda} rather than a \acro{ch} for faster transactions and higher volumes.

One avenue of future research is to combine this evaluation method with evolutionary computation to optimize \acro{da} mechanisms.

%---------------------------------------------------------------------
\subsection{Adaptive auction mechanisms}
\label{sec:mechdes:adaptive}
%---------------------------------------------------------------------

Considering that the information about the population of traders is usually unknown to the auction mechanism, and many analytic methods depend on specific assumptions about traders, Pardoe and Stone advocated a self-adapting auction mechanism that adjusts auction parameters in response to past auction results \cite{pardoe05developing-adaptive-mech}.

Their framework includes an \emph{evaluator} module, which can create an auction mechanism for online use, can monitor the performance of the mechanism, and can use the economic properties of the mechanism as feedback to guide the discovery of better parameter combinations. This process then creates better auction mechanisms that continue to interact with traders which are themselves possibly evolving at the same time. A classic algorithm for $n$-armed bandit problems, $\epsilon$-greedy, is used in the evaluator module to make decisions on parameter value selection.

This work differs from previous work in the sense that here auction mechanisms are optimized during their operation while the mechanisms in the approaches discussed before find remain static and are assumed to perform well even when they face a set of traders that is different from those used in searching for the mechanisms.

%---------------------------------------------------------------------
\subsection{Auction mechanism design competition}
\label{sec:mechdes:cat}
%---------------------------------------------------------------------

Following the \acro{tac} \acro{c}lassic and the \acro{tac scm} competitions introduced in Section~\ref{sec:trading:tac}, a new competition called \acro{tac cat}\footnote{\acro{cat} is not only the reverse of \acro{tac}, but also refers to \emph{catallactics}, the science of exchanges.} was run in the summer of 2007 in order to foster research on auction mechanism design. In \acro{tac cat}, the software trading agents are created by the organizers of the competition, and entrants compete by defining rules for matching buyers and sellers and setting commission fees for providing this service. Entrants compete against each other in attracting buyers and sellers and making profits. This is achieved by having effective matching rules and setting appropriate fees that are a good trade-off between making profit and attracting traders.

We developed \acro{jcat} \cite{jcat-manual}, based on Phelps's \acro{jasa},\footnote{\acro{jasa} is a high-performance auction simulator that allows researchers in agent-based computational economics to run trading simulations using a number of different auction mechanisms. The software includes an implementation of the 4-heap algorithm in \cite{wurman98@flexible-double-auction-4-ecommerce} and is designed to be highly extensible, so that new auction rules can easily be implemented. The software also provides base classes for implementing simple adaptive trading agents \cite{jasa}.} to run as the game server. It provides various trading strategies, market selection strategies, and \acro{da} market mechanism frameworks to avoid entrants working from scratch. \acro{jcat} is also an ideal experimental platform for researchers to evaluate auction mechanisms in a competition setting.

%\clearpage

%---------------------------------------------------------------------
\section{Summary}
\label{sec:summary}
%---------------------------------------------------------------------

This report aims to provide an overview of the field of auction mechanism design and build the foundation for further research.

Auctions are markets with strict regulations where traders negotiate and make deals. An auction may be single-sided or double-sided depending upon whether only sellers or only buyers can make offers or whether both can. The four standard single-sided auctions---English auction, Dutch auction, first- and second-price sealed-bid auctions---have been the subject of traditional auction theory. Vickrey's pioneering work in this area led to the revenue equilibrium theorem that shows a seller can expect equal profits on average from all the standard types of auctions with a few assumptions about the bidders. Other researchers followed the approach and managed to extend the applicability of the theorem when the assumptions are relaxed.

Double-sided auctions, which are important in the business world, posed a bigger challenge due to the higher complexity of their structure and the interaction between traders. While classical mathematical approaches have continued to be successful in analyzing some simple types of double auctions, they have been unable to apply to more practical scenarios. Smith and others initiated experimental approaches and showed that double auctions, even with a handful of traders, may lead to high allocative efficiency and the transaction prices quickly converge to the expected equilibrium price. Subsequent experiments with human and/or artificial traders tried to explain what led to these desirable properties and tended to show that auction mechanisms played a major role, though the intelligence of traders had an effect as well.

Further work, on the one hand, introduced more and more complex trading strategies not only making higher individual profits but also improving the collective properties of auctions. On the other hand, different methods have been explored to design novel auction mechanisms. One approach is to evolve parameterized auction mechanisms based on evolutionary computation. Cliff \textit{et al.} have found a new variant of continuous double auctions through evolving mechanisms that converge more quickly to equilibrium, and also exhibit higher efficiency than those previously known. Phelps \textit{et al.} have explored the use of genetic programming and justified the traditional mid-point transaction pricing rule as optimizing efficiency while balancing trader market power. In addition to these off-line techniques for optimization through evolutionary computing, online approaches have been proposed to produce adaptive auction mechanisms, which, with dynamic trader populations, can continuously monitor and improve their performance.

With the understanding of this prior research work, what can be done further at the interface of computer science and economics include: obtaining more insights into double-sided auction mechanisms, inventing novel auction rules, and searching for optimal combinations of various kinds of policies, automatically producing desirable auction mechanisms.

%\clearpage

%%%%%%%%%%%%%%%%%%%%%%%%%%%%%%%%%%%%%%%%%%%%%%%%%%%%%%%%%%%%%%%%%%%%%

\bibliographystyle{plain}

%%%%%%%%%%%%%%%%%%%%%%%%%%%%%%%%%%%%%%%%%%%%%%%%%%%%%%%%%%%%%%%%%%%%%

\end{document}